\documentclass[10pt,twocolumn,letterpaper]{article}

\usepackage[pagenumbers]{cvpr}

\newcommand{\datasetName}{4DP-QA\xspace}

\usepackage{multirow}
\usepackage{multicol}
\usepackage{colortbl}
\usepackage{arydshln}
\usepackage{lipsum}
\usepackage{tikz}
\usepackage{amsmath,bm}
\usepackage{relsize}
\usetikzlibrary{positioning}

\usepackage{makecell}

\usepackage[most]{tcolorbox}

\usepackage{longtable}
\usepackage{pifont}

\newcommand{\mat}[1]{\mathlarger{\mathtt{#1}}}
\newcommand{\vect}[1]{\mathbf{#1}}

\definecolor{cvprblue}{rgb}{0.21,0.49,0.74}
\usepackage[pagebackref,breaklinks,colorlinks,allcolors=cvprblue]{hyperref}

\definecolor{darkgreen}{RGB}{30,150,30}
\definecolor{darkblue}{RGB}{0,0,127}
\definecolor{darkyellow}{RGB}{171,133,0}
\definecolor{darkred}{RGB}{180,20,20}
\definecolor{darkmagenta}{RGB}{127,0,127}
\definecolor{darkcyan}{RGB}{0,127,127}
\definecolor{chromeyellow}{rgb}{1.0, 0.65, 0.0}
\definecolor{amber}{rgb}{1.0, 0.75, 0.0}

\newif\ifdrafting  
\draftingfalse %
\ifdrafting
  \newcommand{\SC} [1] {\textcolor{amber}{[SC: #1]}}
  \newcommand{\OG} [1] {\textcolor{darkgreen}{[OG: #1]}}
  \newcommand{\AB} [1] {\textcolor{darkyellow}{[AB: #1]}}
  \newcommand{\HS} [1] {\textcolor{darkmagenta}{[HS: #1]}}
  \newcommand{\JJ} [1] {\textcolor{darkblue}{[JJ: #1]}}
  \newcommand{\SL} [1] {\textcolor{darkcyan}{[SL: #1]}}
  \newcommand{\ZZ} [1] {\textcolor{darkred}{[ZZ: #1]}}

\else
  \newcommand{\SC} [1] {}
  \newcommand{\OG} [1] {}
  \newcommand{\AB} [1] {}
  \newcommand{\HS} [1] {}
  \newcommand{\JJ} [1] {}
  \newcommand{\SL} [1] {}
  \newcommand{\ZZ} [1] {}

\fi

\def\paperID{945} %
\def\confName{CVPR}
\def\confYear{2026}

\title{4DP-QA: Scalable QA for 4D Perception in Vision Language Models}

\author{Seokju Cho$^{1,3*}$ \quad Abhishek Badki$^{1}$ \quad Hang Su$^{1}$ \quad Jindong Jiang$^{1}$ \quad Ziyao Zeng$^{1,2*}$ \\ \quad Seungryong Kim$^{3}$ \quad Sifei Liu$^{1}$ \quad Orazio Gallo$^{1}$ \\[1em]
$^{1}$NVIDIA \quad $^{2}$Yale University \quad $^{3}$KAIST AI 
}

\begin{document}

\maketitle

\begingroup
\renewcommand\thefootnote{}
\footnote{$^{*}$Work done during internship at NVIDIA.}
\addtocounter{footnote}{-1}
\endgroup

\begin{abstract}
Despite recent advances, Vision Language Models (VLMs) still struggle to grasp the dynamics of the world.
We note that the ability to reason about a 4D scene, challenging in itself, is further complicated by two factors.
First, VLMs observe motion indirectly via its projection onto 2D images.
Second, existing datasets fail to disentangle object and camera motion.
To address these challenges, we present a QA generation pipeline that focuses on motion-related scene understanding.
We take particular care of the entanglement of camera and object motion by casting tracking in both the traditional way and in a novel, fixed reference system, dubbed \textit{True-Motion Tracking}, which provides an intuitive description of motion.
From this pipeline, we generate a large-scale training dataset of 400K samples, \datasetName (4D Perception QA), and a 2.2K-sample benchmark, \datasetName-Bench.
Training existing models on our dataset yields performance improvements on an external benchmark, validating the effectiveness of our method.
\end{abstract}

\begin{figure}[t]
    \centering
    \includegraphics[width=\columnwidth]{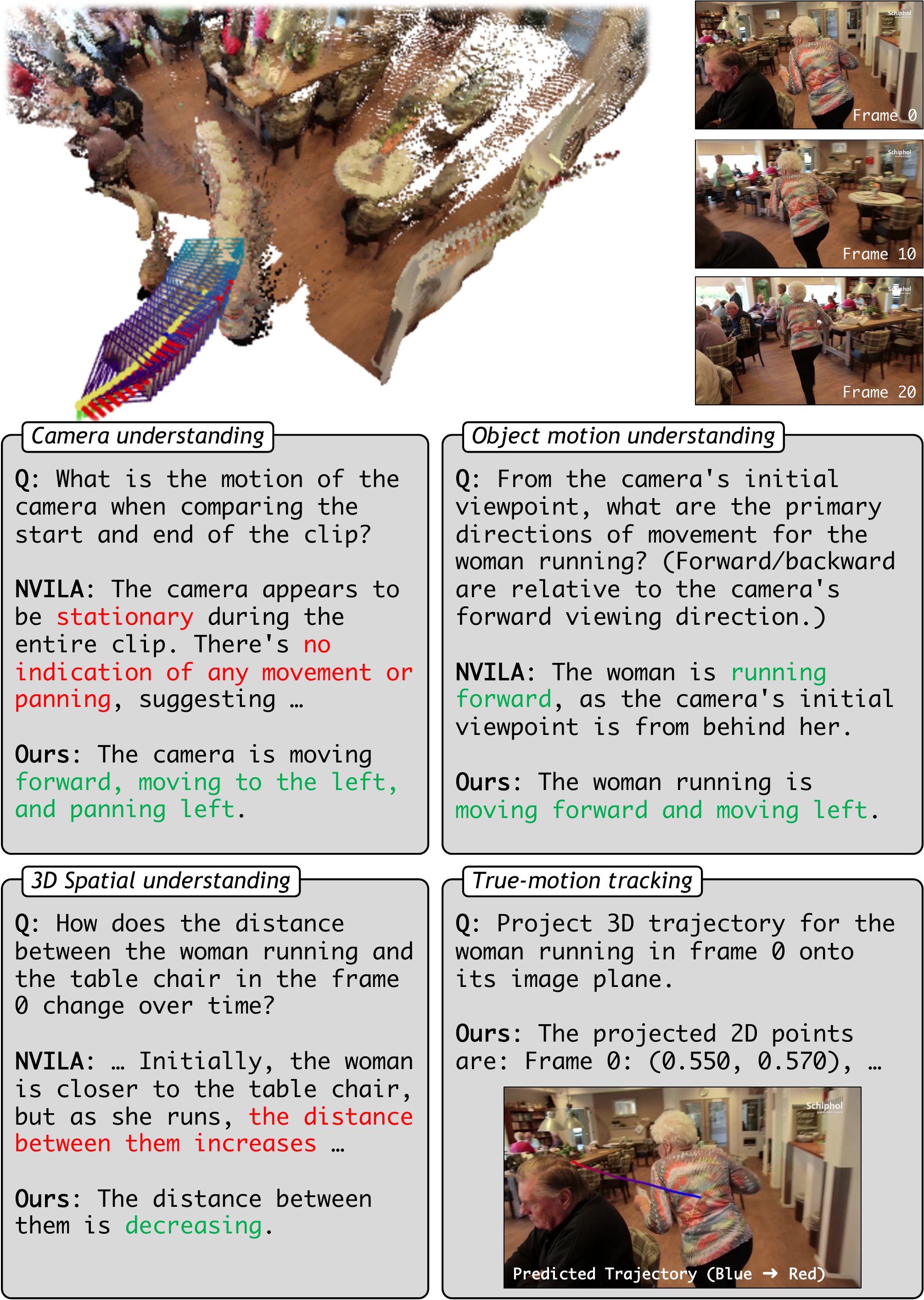}
    \captionsetup{hypcap=false}
    \caption{
        Our framework equips VLMs with better 4D understanding for in-the-wild videos. 
        Training a state-of-the-art VLM (NVILA~\cite{liu2024nvila}) on our dataset yields performance gains (NVILA vs.~Ours).
        We also introduce \textit{true-motion point tracking}, a new capability that enables the VLM to isolate true object motion from camera movement, leading to better 4D understanding.}
    \label{fig:teaser}
    \vspace{-1em}
\end{figure}

\section{Introduction}

Vision Language Models (VLMs)~\cite{zhang2024vlmsurvey} have made great progress in semantic reasoning with images and videos~\cite{liu2023visual, liu2024nvila, bai2025qwen2, wang2024internvideo2, chen2025eagle2.5}, and 3D scene understanding~\cite{chen2024spatialvlm, azuma2022scanqa, ma2022sqa3d, chen2021scan2cap}. 
However, their grasp of motion, ubiquitous in the physical world, remains limited.
This is in part rooted in the video capture process itself: the world is 4D (3D+motion), but it is projected onto the 2D sensor of a camera, which is generally also moving.
As a result, depth cues are discarded and, even more critically, the absolute motion of objects is entangled with that of the camera, making 4D understanding challenging.
Compounding this, most training datasets focus on understanding the apparent motion (motion in the 2D image plane), rather than the true 3D motion.
In this work, we address this limitation by introducing a comprehensive framework to equip VLMs with better 4D understanding.

Large models exhibit strong \emph{scaling properties}~\cite{kaplan2020scalingllm,radford2021learning}, which allow the \emph{emergence} of new capabilities.
Our intuition is that 4D understanding will similarly emerge, even in largely standard VLM architectures, given the right \emph{quality} and \emph{quantity} of training data.
To achieve this, we introduce a scalable spatio-temporal QA generation pipeline that systematically analyzes scene dynamics using accurate geometric information (camera poses, depths, 6D object poses, \etc) from multiple data sources, both real and synthetic.
We use this pipeline to construct QA pairs that probe different aspects of scene dynamics, such as camera motion, object motion, inter-object dynamics, and spatial reasoning. 
Our approach employs carefully designed heuristics to translate continuous geometric measurements into natural language descriptions tailored to the different question types. 

\begin{figure}[t]
    \centering
    \includegraphics[height=1.5in]{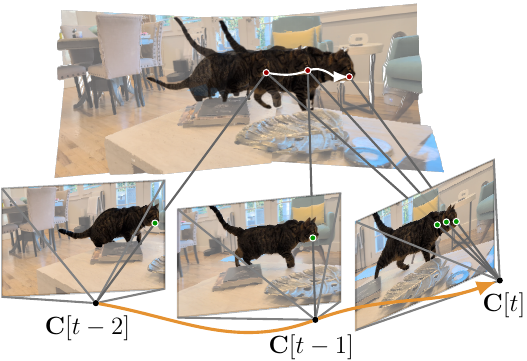}\\
    \vspace{2mm}
    \subfloat[][Visual Point Tracking]{\includegraphics[width=.45\columnwidth]{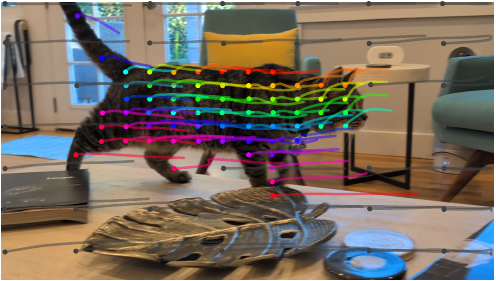}}
    \subfloat[][True-motion Point Tracking]{\includegraphics[width=.45\columnwidth]{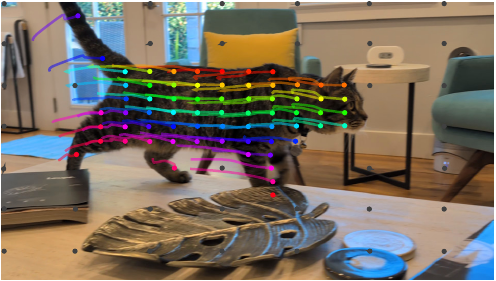}}
    \caption{\textbf{True-motion Point Tracking.} Visual point tracking (a) only captures the apparent motion of the object, here making the cat appear to move backward due to the rightward camera motion. True-motion point tracking (b) disentangles camera and object motion, showing the cat moving forward. Background tracks (gray points) show movement in (a) but remain stationary in (b), highlighting that true-motion tracks reflect actual object motion as seen from a fixed viewpoint.}
    \label{fig:truemotion}
\end{figure}

While these natural language QA pairs provide rich semantic supervision, coarse language descriptions alone are insufficient for fine-grained 4D understanding. %
To address this, we develop QA pairs aimed at more fine-grained and low-level perceptual abilities.
One such task is \emph{visual point tracking}, that is, following points~\cite{doersch2022tap,karaev2024cotracker} across video frames, crucial for a range of applications like 3D reconstruction~\cite{wang2023vggsfm} and robotics~\cite{wen2024anypointtrajectory,bharadhwaj2024track2act}.
Visual point tracking allows us to visually associate dynamic regions across frames, but the tracks do not always give a meaningful signal about the \emph{true motion} of the object when the camera is moving.
For instance, in Figure~\ref{fig:truemotion} the camera is moving to the right faster than the cat. As a result, the visual point tracks (Figure~\ref{fig:truemotion}(a)) show points on the cat moving backward.

To address this, we introduce a new perceptual task: estimating an object's motion as if it were observed from a stationary reference system, such as the viewpoint of a static camera.
Under this formulation, the object's motion is disentangled from the camera's motion, providing a more intuitive representation: now the tracks show the cat moving forward, see Figure~\ref{fig:truemotion}(b).  
We term this capability \emph{true-motion point tracking}.
Visual point tracking encourages the model to capture dense motion correspondences tied to appearance changes, while true-motion point tracking teaches it to reason about object motion in a stable, fixed reference system.
We incorporate these fundamental perceptual tasks into our dataset.

Our work makes the following contributions:
\begin{itemize}
    \item We introduce a scalable spatio-temporal QA generation pipeline that automatically produces high-quality 4D reasoning data, yielding a training dataset with 400K QA pairs (\datasetName), and a benchmark with 2.2K QA pairs (\datasetName-Bench).
    \item We train multiple VLMs on our training dataset and show improved performance on both ours and external 4D reasoning benchmarks.
    \item We introduce true-motion point tracking, a new low-level perceptual task that provides an intuitive, image-aligned motion representation. To enable VLMs to learn this task, we develop dedicated QA pairs covering both true-motion and visual point tracking.
    \item We show that incorporating the point tracking tasks into our training data enhances the spatio-temporal reasoning abilities of VLMs on external benchmarks, underscoring the value of fine-grained perceptual supervision.
\end{itemize}

\section{Related Work}

This work investigates architectural and data strategies to equip VLMs with 3D perception and motion understanding, including robust capabilities for visual and motion-based point tracking. 
In the following, we review previous work on relevant models, datasets, and point tracking.

\vspace{-1em}\paragraph{General VLMs.} 
The foundation for 3D/4D VLMs rests on 2D models. 
Early approaches like Flamingo~\cite{alayrac2022flamingo} and BLIP-2~\cite{li2023blip} focused on aligning visual encoders with LLMs. 
Subsequent work moved toward instruction tuning~\cite{liu2023visual} and scaling~\cite{lin2024vila, chen2025eagle2.5, chen2024internvl, bai2025qwen2}.
Video VLMs extend 2D VLMs to temporal understanding, with rapid advances in training strategies~\cite{wang2024internvideo2}, computational efficiency~\cite{jiang2025token, li2025frameoracle}, and long-range reasoning~\cite{xie2025video, hannan2025revisionllm}. 
Research is also converging on integrated 4D models~\cite{zheng2025video, zhou2025uni4d} that fuse video with 3D awareness for physical-world reasoning.

\vspace{-1em}\paragraph{3D VLMs.}
3D VLMs aim to equip LLMs with 3D spatial understanding. 
We focus on scene-level VLMs rather than those operating on raw 3D data like point clouds~\cite{xu2024pointllm, qi2024gpt4point} or meshes~\cite{fang2025meshllm, wang2024llama}, which primarily address shape-level geometry and not complex, holistic scene reasoning.
These approaches often integrate 3D information via alignment, such as depth map projection~\cite{cheng2024spatialrgpt} and 3D position embeddings~\cite{cheng20253d, zhu2024llava}.
VLM-3R~\cite{fan2025vlm} bypasses the reliance on explicit 3D data and adopts CUT3R~\cite{wang2025continuous} for 3D reconstructive tokenization. 
SpatialVLM~\cite{chen2024spatialvlm} demonstrates that quantitative spatial reasoning is achievable with only 2D inputs, enabled by a massive, synthesized 2B-sample VQA dataset.

\vspace{-1em}\paragraph{Spatial-temporal understanding datasets.}
Compared to datasets designed for general image and video understanding~\cite{liu2024mmbench, li2024seed} or large-scale vision-language instruction tuning~\cite{li2024llava, lin2024vila}, datasets available for training and benchmarking VLMs' spatial-temporal capabilities are substantially more limited in scope or scale. 
Static 3D datasets~\cite{azuma2022scanqa, ma2022sqa3d, chen2021scan2cap, zhang2025flatland} lack dynamics beyond passive camera motion. 
General video datasets~\cite{fu2025video, yang2025thinking, wang2025spatialvid} address scene dynamics and temporal reasoning but neglect explicit 3D geometry and true-motion disentanglement necessary for robust 4D perception. 
Works targeting dynamic scenes with 3D annotations are either limited to benchmarking scale~\cite{zhou2025vlm4d, li2025sti}, restricted in temporal complexity~\cite{ray2024sat}, or constrained to limited contexts like automotive scenes~\cite{tian2024drivevlm}.
ST-VLM~\cite{ko2025st}, a concurrent work, focuses on general scenes but is limited in scale compared to our work.
We propose a data pipeline to create a large-scale dataset of fine-grained spatial-temporal QA pairs to overcome these limitations.

\vspace{-1em}\paragraph{Point Tracking.}
Point tracking has evolved from 2D correspondences~\cite{doersch2022tap, doersch2023tapir, cho2024local, karaev2024cotracker} to 3D tracking~\cite{xiao2025spatialtracker, cho2025seurat} and more recently towards full 4D understanding with joint modeling of camera and object dynamics~\cite{tapip3d, xiao2025spatialtrackerv2}.
While 2D methods capture apparent motion and 3D methods add depth awareness, only 4D approaches aim to recover true object motion by disentangling camera and object motion.
However, these solutions are highly specialized, making it challenging to integrate them into general-purpose VLMs.

\section{4D Understanding with VLMs: A Data-Driven Approach}

We present a data-driven approach to equip VLMs with fine-grained 4D perception capabilities.
We first introduce the perceptual task of true-motion point tracking (Section~\ref{subsec:tracking}), which later serves as a key component in defining novel QA pairs. 
We then introduce a scalable QA generation pipeline (Section~\ref{subsec:pipeline}), 
with which we create a large-scale 4D understanding QA dataset (Section~\ref{subsec:datasets}) for training and evaluating VLMs.

\subsection{Preliminaries: True-Motion Point Tracking}
\label{subsec:tracking}

For a 3D point track $\{\vect{X}[t]\}$ over a time window $[0, T)$ and a moving camera with extrinsics $\{\displaystyle\mat{T}[t]\}$ and intrinsics $\displaystyle\mat{K}$, there are two ways to image the 3D point track in 2D:
\begin{equation}
    \label{eq:visual_point_track}
    P_{2D} = \{\vect{p}[t]\}_{t \in [0, T)} = \{\Pi(\mat{K}, \mat{T}[t], \vect{X}(t))\}_{t \in [0, T)},
\end{equation}
\begin{equation}
    \label{eq:true_motion_track}
    M_{2D} = \{\vect{m}_{t_q}[t]\}_{t \in [0, T)} = \{\Pi(\mat{K}, \mat{T}[t_q], \vect{X}(t))\}_{t \in [0, T)},
\end{equation}
where $\Pi$ is the image projection operator, and $t_{q}$ is a fixed reference time.
Equation~\ref{eq:visual_point_track} gives us the 2D visual point track $P_{2D}$, while Equation~\ref{eq:true_motion_track} defines what we term the \emph{true-motion} point track $M_{2D}$.
The difference between them is that $P_{2D}$ is imaged by a changing camera, thus entangling the object and the camera motion, while $M_{2D}$ is imaged by a fixed reference system of the camera at time $t_q$, thus better capturing the true motion of the object.
They become equivalent when the camera is stationary, since $\mat{T}[t] = \mat{T}[t_q]$.

Given a 2D query pixel $\vect{p}[t_q]$ at time $t_q$ (where $\vect{p}[t_q] = \vect{m}_{t_q}[t_q]$), the tasks of visual point tracking and true-motion point tracking are to predict the whole sequences $P_{2D}$ and $M_{2D}$, respectively.
Visual point tracking can be solved using correspondences by comparing the features of $\vect{p}[t_q]$ with pixels at other time in the image sequence~\cite{doersch2022tap}.
On the other hand, to estimate the true-motion point track, one needs to not only solve the visual correspondence task but also recover depth and camera poses.
In return, the true-motion track gives an intuitive image-aligned representation that allows reasoning about the dynamic motion in context of its static surrounding from a fixed desired reference system, as shown in Figure~\ref{fig:truemotion}.
We argue that the two are complementary and both are useful for understanding scene dynamics. 

Visual point tracking is a well-studied task with specialized solutions~\cite{cho2024local,karaev2024cotracker}. 
Likewise, true-motion point tracks can be estimated with modern 3D reconstruction methods by explicitly reconstructing 3D point tracks~\cite{xiao2025spatialtrackerv2,tapip3d,feng2025st4rtrack,badki2025l4p}.
Integrating these solutions into VLMs requires careful design and brings additional overhead and complexity.
Instead, we formulate both tracking tasks as QA pairs and design a large-scale QA generation pipeline to investigate VLMs' capabilities to solve these tasks directly.

\begin{table}[t]
    \centering
    \caption{\textbf{Data sources for the dataset generation pipeline.} We collect data from a variety of sources, including driving, indoor, and simulation datasets, spanning synthetic and real-world scenes.
    Once preprocessed, they are standardized to a common format that our generation pipeline (Section~\ref{subsec:pipeline}) can use to produce QA pairs.}
    \label{tab:dataset_sources}
    \resizebox{\linewidth}{!}{%
    \begin{tabular}{@{}l l l r r@{}}
    \toprule
    \multirow{2}{*}{Dataset} & \multirow{2}{*}{Type} & \multirow{2}{*}{Domain} & \multicolumn{2}{c}{Number of} \\
    \cmidrule(l){4-5}
    & & & Videos & Frames \\
    \midrule
    SHIFT~\cite{sun2022shift} & Driving & Synthetic & 3,200 & 1.6M\\
    Virtual KITTI 2~\cite{cabon2020virtual} & Driving & Synthetic & 50 & 21.3K\\
    Aria Digital Twin~\cite{pan2023aria} & Indoor, Egocentric & Real & 273 & 644K\\
    HOT3D~\cite{banerjee2024introducing} & Indoor, Egocentric & Real & 424 & 1M\\
    Kubric~\cite{greff2022kubric} & Simulation & Synthetic & 9.7K & 320K\\
    \midrule
    \multicolumn{3}{l}{Total}&13.6K&\textbf{3.6M}\\
    \bottomrule
    \end{tabular}
    }
\end{table}

\begin{figure*}[t]
    \centering
    \includegraphics[width=\linewidth]{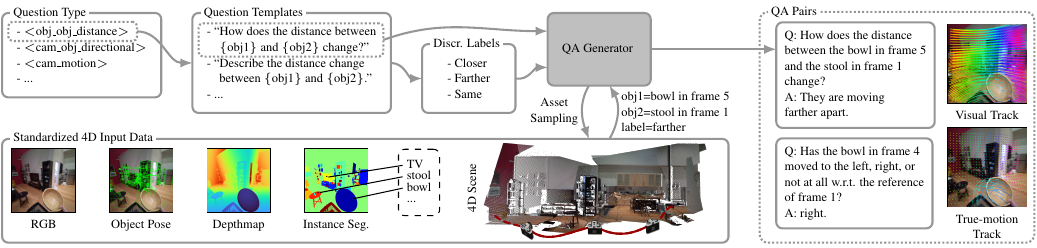}
    \caption{\textbf{Dataset Generation Pipeline (Section~\ref{subsec:pipeline}).} The pipeline takes as input standardized 4D input data, and produces QA pairs for 13 question types. 
    Each QA pair is generated by instantiating a pre-defined template with the sampled assets and either their discrete labels (for descriptive questions) or continuous measurements (for visual and true-motion point tracking). 
    }
    \label{fig:dataset_pipeline}
\end{figure*}

\subsection{Dataset Generation Pipeline}
\label{subsec:pipeline}
We design a scalable pipeline to generate 3D motion QA pairs from a wide range of data sources, including automotive, physics simulation, and indoor datasets featuring human-object interactions (Table~\ref{tab:dataset_sources}).
To be consumed by the generation pipeline, each data source is first standardized to a common format of coordinates, image resolution, frame rates, camera parameters, depth, segmentation, metadata, and 6D object poses.
3D point tracks required for many of the QA pairs are extracted from the provided depth maps, camera poses, and 6D object poses.
Object labels, when available, were insufficient to uniquely identify objects, so we use a VLM~\cite{bai2025qwen2} to caption each object using its bounding box and initial object label. 
After preprocessing, our QA generation pipeline proceeds through the components outlined below.
See Figure~\ref{fig:dataset_pipeline} for an overview. 
Our pipeline is readily extendable to arbitrary video sources by utilizing off-the-shelf 4D reconstruction methods~\cite{lin2025depth,wang2025continuous}.

\vspace{-1em}\paragraph{QA generator.}
This is the core component that produces QA pairs for each question type. 
It first initializes QA pairs by sampling from a set of pre-defined templates.
We design multiple templates for each question type and use an off-the-shelf LLM (Gemini-2.5-Pro~\cite{comanici2025gemini2}) to produce diverse phrasing for both questions and answers.
Each template contains slots for object references, time instances, coordinate systems, as well as slots in the answers for discrete labels (for descriptive questions) or continuous measurements (for visual and true-motion point tracking).
The QA generator then informs the asset sampling process about certain desired characteristics of the assets (\eg, there should be $\geq 2$ moving objects) for the sampled template. 
Last, we finalize QA pairs by filling in the slots with the returned assets that meet the requirements.

\vspace{-1em}\paragraph{Assets sampling.}
Not all video segments display meaningful or sufficient motion for the targeted question types.
We first sample random video segments, then assess each segment for suitability according to the specified requirements.
To do so, we design a set of heuristics and dataset-specific thresholds to determine whether the camera and objects display the desired characteristics.
Depending on the question, it may be necessary for the camera to remain stationary or move in a particular manner. 
Similarly, objects are filtered based on their motion characteristics (\eg, stationary, moving, changing direction) and the degree to which they are visible within the video segment.
We then sample from this filtered object list, discarding the segment if there are insufficient eligible objects for the target question.
For certain question types, we also need to choose the appropriate reference system, such as a specific camera coordinate system, the object's canonical reference system or screen-space coordinate system, in order to identify the selected objects.
Finally, we sample how to refer to the objects in the questions: by object caption, by coordinate (pixel \((x,y)\) in a specific frame), or by visual region annotation (circle around the region of interest~\cite{shtedritski2023does}).

\vspace{-1em}\paragraph{Discrete label generation.}
For question types requiring coarse descriptive answers, care needs to be taken to map the continuous geometry measurements like distances, translations, rotations, directions, \etc computed in a specific reference system into categorical labels.
For example, 3D translation may be described using combinations of (forward/backward, left/right, up/down), 3D rotation using combinations of (pan left/right, tilt up/down, roll clockwise/counterclockwise), 3D distances using categories (increasing/decreasing/constant); similar descriptors are used for 2D screen-space motions (see supplementary).
We carefully designed the thresholds for each data source to ensure that the selected labels for questions are accurate, unambiguous, and aligned with human perceptual judgments.
For questions surrounding the perceptual capabilities of visual and true-motion point tracking, we directly pass the continuous values to the QA generator.

Finally, we perform multiple rounds of human validation to ensure high quality for the QA pairs in our benchmark.

\subsection{The \datasetName Dataset}
\label{subsec:datasets}

With the scalable QA generation pipeline, we construct our dataset, \datasetName, with the goal of equipping standard VLMs~\cite{liu2024nvila,bai2025qwen2} with fine-grained 3D motion understanding across a variety of scenes.
The QA pairs cover perceptual tasks like visual and true-motion point tracking as well as a broad array of 3D motion understanding challenges—such as camera motion, object motion, and spatial reasoning.
\textbf{Key characteristics} of \datasetName include: (1)~consistent use of relative scales to support generalization across diverse scenes, (2)~multiple modes of object reference (visual prompts, coordinates, and captions), (3)~deliberate use of different reference systems (including camera, object, screen-space, and gravity-aligned) in formulating questions, 
(4)~varied answer formats (free-form descriptive, multiple-choice, and Y/N responses). 

\datasetName includes 13 question types that can be organized into 4 categories as described below.
We describe the high-level dataset taxonomy in this section and provide additional details in Section~\ref{subsec:dataset_details} and the supplementary. 

\vspace{-1em}\paragraph{I.~Camera Motion.} 
This category concerns the motion of the camera, as well as how the camera's motion affects its relation to objects.
Question types include:
(1)~\emph{Camera Movement}: how the camera translates and rotates over time.
(2)~\emph{Camera-Object Distance Change}: how the distance between the camera and an object changes. %

\vspace{-1em}\paragraph{II.~Object Motion.} 
This category concerns the motion of objects relative to the camera or other objects.
Question types include:
(3)~\emph{Rotation}: how an object rotates from a top-down view.
(4)~\emph{Direction}: how an object translates relative to the first frame's viewpoint.
(5)~\emph{Agent Motion}: how an agent (\eg, person or car) moves \wrt its own reference system.
(6)~\emph{Moved Distance}: comparing the distances traveled by two objects.
(7)~\emph{Object-Object Distance Change}: how the distance between two objects changes over time.

\vspace{-1em}\paragraph{III.~3D Spatial Understanding.} 
This category concerns the understanding of 3D spatial relationships between objects as well as with respect to the camera.
The multi-view questions (10 and 11) introduce further complexity, requiring joint reasoning about object motion and depth variation across time.
Question types include:
(8)~\emph{Depth}: comparing the distances of objects from the camera.
(9)~\emph{Object-Object Distance}: comparing the distances of objects from a reference object within the same frame.
(10)~\emph{Multi-View Depth}: similar to 8, but with reference to objects in different frames.
(11)~\emph{Multi-View Object-Object Distance}: similar to 9, but with reference to objects in different frames.

\vspace{-1em}\paragraph{IV.~Point Tracking.} These questions require recovering the tracks of (12)~\emph{Visual Point Tracking}, $P_{2D}$ (Equation~\ref{eq:visual_point_track}), and (13)~\emph{True-Motion Point Tracking}, $M_{2D}$ (Equation~\ref{eq:true_motion_track}), given a query point in a reference frame. 
In the question, we also specify the target frames (which can be a single or multiple frames) for which we need to recover the track. 
The answer includes the coordinates and an occlusion flag indicating the visibility of the point in each target frame.

\begin{table*}[t]
    \centering
    \caption{\textbf{Quantitative comparison of VLMs on our \datasetName-Bench.} We report the accuracy (\%) for each task. Alongside comparisons with off-the-shelf VLMs, we also present the performance improvements obtained by training the baseline models on \datasetName. Please refer to Section~\ref{subsec:datasets} for detailed descriptions of each column.}
    \label{tab:main_quantitative}
    \resizebox{0.95\textwidth}{!}{%
    \begin{tabular}{@{}l *{15}{w{c}{2.6em}}@{}}
    \toprule
    & \multicolumn{3}{c}{\textbf{Camera Motion}} & \multicolumn{6}{c}{\textbf{Object Motion}} & \multicolumn{5}{c}{\textbf{3D Spatial Understanding}} & \\
    \cmidrule(lr){2-4} \cmidrule(lr){5-10} \cmidrule(lr){11-15}
    \textbf{Model} & 
    \rotatebox{90}{(1) Cam.} & \rotatebox{90}{(2) Cam.-Obj. Chg.} & \rotatebox{90}{\textbf{Average}} &
    \rotatebox{90}{(3) Rotation} & \rotatebox{90}{(4) Direction} & \rotatebox{90}{(5) Agent} & \rotatebox{90}{(6) Moved Dist.} & \rotatebox{90}{(7) Obj.-Obj. Chg.} & \rotatebox{90}{\textbf{Average}} &
    \rotatebox{90}{(8) Depth} & \rotatebox{90}{(9) Obj.-Obj. Dist.} & \rotatebox{90}{(10) MV Depth} & \rotatebox{90}{(11) MV Obj. Dist.} & \rotatebox{90}{\textbf{Average}} &
    \rotatebox{90}{\textbf{Overall Avg.}} \\
    \midrule
    Random & 43.7 & 39.2 & 41.5 & 50.0 & 25.0 & 50.0 & 32.0 & 31.7 & 28.5 & 50.0 & 50.0 & 50.0 & 50.0 & 50.0 & 40.8 \\
    \midrule
    \multicolumn{16}{@{}l}{\textit{Proprietary VLMs}} \\
    GTP-4o~\cite{hurst2024gpt4o} & 53.3 & 51.0 & 52.1 & 51.3 & 40.5 & 18.2 & 44.3 & 32.3 & 41.7 & 67.1 & 67.8 & 64.2 & 58.2 & 65.2 & 53.8 \\
    Gemini-2.5-pro~\cite{comanici2025gemini2} & 58.4 & 67.5 & 63.2 & 53.1 & 46.1 & 45.5 & 55.7 & 56.9 & 50.8 & 82.6 & 83.1 & 81.8 & 80.6 & 82.2 & 66.8 \\
    Gemini-2.5-flash~\cite{comanici2025gemini2} & 49.6 & 54.0 & 51.6 & 55.8 & 37.1 & 18.2 & 57.1 & 50.8 & 46.8 & 78.8 & 82.5 & 80.4 & 74.6 & 79.3 & 60.5 \\
    Gemini-robotics~\cite{team2025geminirobotics} & 52.8 & 51.6 & 52.4 & 54.0 & 41.4 & 36.4 & 59.3 & 51.5 & 48.8 & 83.2 & \cellcolor{red!20}86.9 & 82.4 & 75.4 & 82.6 & 62.5 \\
    \midrule
    \multicolumn{16}{@{}l}{\textit{Open-source VLMs}} \\
    Qwen2-VL-7B~\cite{bai2025qwen2} & 48.2 & 38.8 & 43.8 & 23.0 & 35.3 & 09.1 & 35.0 & 19.2 & 31.1 & 56.5 & 59.6 & 52.7 & 52.2 & 55.8 & 44.2 \\
    Qwen3-VL-8B~\cite{yang2025qwen3} & 54.5 & 42.1 & 48.7 & 50.4 & 32.3 & 27.3 & 43.6 & 33.8 & 39.6 & 54.7 & 53.0 & 46.6 & 51.5 & 52.3 & 47.3 \\

    Qwen2.5-VL-32B~\cite{bai2025qwen2} & 52.8 & 52.8 & 52.9 & 40.7 & 44.0 & 27.3 & 40.0 & 38.5 & 42.9 & 63.2 & 69.4 & 55.4 & 52.2 & 61.4 & 52.9 \\
    Qwen3-VL-32B~\cite{yang2025qwen3} & 60.6 & 53.7 & 58.2 & 62.8 & 43.5 & 36.4 & 33.6 & 38.5 & 46.9 & 53.2 & 50.3 & 52.0 & 50.0 & 51.8 & 52.1 \\
    \arrayrulecolor{lightgray}\hdashline
    LLAVA-OneVision-7B~\cite{li2024llava} & 47.7 & 49.6 & 48.7 & 34.5 & 38.8 & 18.2 & 33.6 & 28.5 & 36.0 & 54.1 & 59.6 & 44.6 & 50.0 & 52.9 & 46.3 \\
    LLAVA-OneVision-1.5-8B~\cite{an2025llavaonevision15} & 46.2 & 33.4 & 40.2 & 38.9 & 40.1 & 36.4 & 35.0 & 32.3 & 39.1 & 60.0 & 65.6 & 51.4 & 59.7 & 59.6 & 46.5 \\
    \arrayrulecolor{lightgray}\hdashline

    NVILA-Video-8B~\cite{liu2024nvila} & 45.3 & 24.2 & 36.8 & 26.5 & 40.5 & 09.1 & 42.9 & 30.0 & 35.6 & 52.6 & 52.5 & 56.1 & 50.7 & 52.9 & 42.2 \\

    \arrayrulecolor{black}\midrule
    \multicolumn{16}{@{}l}{\textit{Baseline vs. Trained on \datasetName (Ours)}} \\

    Qwen2.5-VL-3B~\cite{bai2025qwen2} & 50.4 & 41.5 & 47.1 & 33.6 & 39.2 & 27.3 & 40.0 & 29.2 & 36.9 & 58.5 & 56.3 & 49.3 & 47.0 & 54.4 & 46.7 \\
    Qwen2.5-VL-3B + \datasetName & 81.0 & 81.5 & 81.3 & 72.6 & 71.6 & \cellcolor{red!20}90.9 & 78.6 & 73.1 & 73.9 & 88.8 & 85.2 & 91.9 & 78.4 & 86.8 & 81.3 \\
    \arrayrulecolor{lightgray}\hdashline
    Qwen2.5-VL-7B~\cite{bai2025qwen2} & 45.0 & 30.7 & 39.5 & 54.9 & 43.5 & 27.3 & 40.0 & 36.9 & 45.1 & 60.3 & 59.6 & 48.0 & 50.0 & 56.1 & 46.6 \\
    Qwen2.5-VL-7B + \datasetName & \cellcolor{red!20}83.7 & \cellcolor{red!20}85.7 & \cellcolor{red!20}84.4 & 75.2 & 81.0 & 72.7 & \cellcolor{red!20}79.3 & \cellcolor{red!20}79.2 & 79.6 & 90.0 & 84.7 & 91.9 & \cellcolor{red!20}83.6 & 88.1 & 84.3 \\

    \arrayrulecolor{lightgray}\hdashline
    NVILA-Lite-8B~\cite{liu2024nvila} & 50.1 & 34.0 & 42.4 & 16.8 & 30.2 & 18.2 & 28.6 & 20.8 & 26.0 & 55.6 & 54.1 & 58.1 & 53.7 & 55.4 & 42.3 \\
    NVILA-Lite-8B + \datasetName & 82.0 & 85.1 & 83.5 & \cellcolor{red!20}76.1 & \cellcolor{red!20}86.6 & 63.6 & 76.4 & 76.9 & \cellcolor{red!20}81.6 & \cellcolor{red!20}90.3 & 86.3 & \cellcolor{red!20}93.9 & 81.3 & \cellcolor{red!20}88.6 & \cellcolor{red!20}84.4 \\
    \arrayrulecolor{black} 
    \bottomrule
    \end{tabular}%
    }
    \end{table*}

\section{Implementation}
\label{sec:implementation}

We present key dataset and training details below, with further information in the supplementary.

\subsection{Dataset Details}
\label{subsec:dataset_details}
Using our proposed dataset pipeline, we construct two datasets: the large-scale \datasetName dataset for training and a standalone \datasetName-Bench dataset for benchmarking. 

\vspace{-1em}\paragraph{\datasetName.} 
For training, we generate 400K QA pairs derived from videos totaling 3.3M frames. 
We sample 32 frames per video segment and use \(448\times448\) resolution.
The dataset contains diverse answer formats: multiple-choice, Y/N, free-form, and point tracking trajectories.
Objects are referenced via visual prompts (circles~\cite{shtedritski2023does}), pixel coordinates, or captions.
Tracking coordinates are normalized to $[0, 1]$ and rounded to three decimals.
To prevent shortcut learning from autoregressive extrapolation, we randomize the order of trajectory frames in outputs and include frame indices in queries to specify which frames to track.

\vspace{-1em}\paragraph{\datasetName-Bench.}
For benchmarking, we generate 2.2K QA pairs from 317K held-out test frames, covering camera motion, object motion, and 3D spatial understanding.
Questions are formatted as either multiple-choice with four options or binary Y/N.
We evaluate model responses using exact string matching following~\cite{yang2025thinking} and report accuracy as the final metric.
We provide additional QA quality and bias analysis in the supplementary.

\subsection{Training Details}
\label{subsec:training_details}

We study two architectures: (1) standard VLMs (NVILA-Lite-8B~\cite{lin2024vila}, Qwen2.5-VL-3B/7B~\cite{bai2025qwen2}) to assess our dataset, and (2) a 4D VLM with an integrated geometry encoder~\cite{badki2025l4p} for injecting pre-trained geometric features, inspired by recent 3D VLMs~\cite{fan2025vlm}.

Standard VLMs are trained with a batch size of 128 for 1 epoch, corresponding to 3.1K training iterations. 
For NVILA-Lite-8B~\cite{liu2024nvila}, we use a learning rate of $2\times10^{-5}$ and AdamW optimizer with a cosine learning rate scheduler. 
Qwen-VL~\cite{bai2025qwen2,yang2025qwen3} variants are trained using the same hyperparameters, except with a lower learning rate of $1\times10^{-5}$.
We freeze the visual encoder during training.
Training of each model takes approximately 9 hours on 32 NVIDIA A100 GPUs with 4-step gradient accumulation.

For our 4D VLM, we integrate L4P~\cite{badki2025l4p}, a general-purpose ViT-based geometry encoder pre-trained on various low-level 4D perception tasks (such as depth and optical flow estimation, and 2D/3D tracking), to inject geometric features directly into the VLM backbone (NVILA-Lite-8B). 
We extend L4P by incorporating the true-motion point tracking task.
L4P features are projected via an MLP and interleaved with visual tokens from the image encoder.
Training follows a two-stage procedure, with visual and geometry encoders frozen throughout.
In the first stage, all model components are frozen except the MLP projector for the geometry encoder, which is trained for 1.5K iterations on 200K randomly sampled QA pairs from the training set.
This stage aligns pre-trained L4P features with the LLM input space.
In the second stage, we unfreeze the LLM and both MLP projectors (for the visual and geometry encoders), and continue training for 3.1K iterations on the full training set in the same manner as standard VLM training.

\section{Experiments}
\label{sec:experiments}

We train baseline VLM models on our dataset (\datasetName) as described in Section~\ref{subsec:training_details} and evaluate them on our benchmark (\datasetName-Bench). 
To test generalization from training on our dataset, we further evaluate on VLM4D~\cite{zhou2025vlm4d}, a 4D reasoning benchmark.
Finally, we conduct ablations to analyze the importance of the point tracking tasks and the integration of a geometry encoder into the VLM.

\subsection{Results on \datasetName-Bench}

Table~\ref{tab:main_quantitative} summarizes model performance on \datasetName-Bench across different question categories.
We evaluate both proprietary models (GPT-4o~\cite{hurst2024gpt4o}, Gemini-2.5-Pro/Flash~\cite{comanici2025gemini2}, Gemini-Robotics~\cite{team2025geminirobotics}) and open-source models (Qwen-VL~\cite{bai2025qwen2,yang2025qwen3}, LLaVA-OneVision~\cite{li2024llava,an2025llavaonevision15}, NVILA~\cite{liu2024nvila}).
We also compare VLM baselines fine-tuned on \datasetName.

\begin{table}[t]
    \centering
    \caption{\textbf{Evaluation on VLM4D~\cite{zhou2025vlm4d} Benchmark.} We evaluate how training on our \datasetName dataset improves generalization to the VLM4D benchmark.} 
    \resizebox{0.85\linewidth}{!}{%
    \begin{tabular}{@{}l >{\centering\arraybackslash}p{1.2cm} >{\centering\arraybackslash}p{1.2cm} >{\centering\arraybackslash}p{1.2cm}@{}}
    \toprule
    \textbf{Model} & \textbf{Real} & \textbf{Synthetic} & \textbf{Overall} \\
    \midrule
        \multicolumn{4}{@{}l}{\textit{Proprietary VLMs}} \\
        Gemini-2.5-Pro~\cite{comanici2025gemini2} & \cellcolor{red!20}62.7 & 62.9 & \cellcolor{orange!20}62.8 \\
        Gemini-2.5-Flash~\cite{comanici2025gemini2} & 51.4 & 50.3 & 51.1 \\
        Gemini-robotics~\cite{team2025geminirobotics} & \cellcolor{orange!20}61.9 & 51.5 & 59.3 \\
        \midrule
        \multicolumn{4}{@{}l}{\textit{Open-source VLMs}} \\
        Qwen2-VL-7B~\cite{bai2025qwen2} & 45.0 & 45.4 & 45.1 \\
        Qwen3-VL-8B~\cite{yang2025qwen3} & 50.6 & 53.9 & 51.4 \\
        Qwen2.5-VL-32B~\cite{bai2025qwen2} & 51.5 & \cellcolor{yellow!20}65.6 & 55.0 \\
        Qwen3-VL-32B~\cite{yang2025qwen3} & 57.0 & 56.0 & 56.8 \\
        \arrayrulecolor{lightgray}\hdashline
        LLAVA-OneVision-7B~\cite{li2024llava} & 46.0 & 33.3 & 42.9 \\
        LLAVA-OneVision-1.5-8B~\cite{an2025llavaonevision15} & 48.1 & 38.9 & 45.9 \\
        \arrayrulecolor{lightgray}\hdashline
        NVILA-Video-8B~\cite{liu2024nvila} & 32.4 & 55.1 & 38.0 \\
        \arrayrulecolor{black}\midrule
        \multicolumn{4}{@{}l}{\textit{Baseline vs. Trained on \datasetName (Ours)}} \\
        Qwen2.5-VL-3B~\cite{bai2025qwen2} & 48.2 & 35.3 & 45.0 \\
        Qwen2.5-VL-3B + \datasetName & 55.0 & 56.9 & 55.5 \\
        \arrayrulecolor{lightgray}\hdashline
        Qwen2.5-VL-7B~\cite{bai2025qwen2} & 52.9 & 50.6 & 52.3 \\
        Qwen2.5-VL-7B + \datasetName & \cellcolor{yellow!20}60.6 & \cellcolor{orange!20}73.0 & \cellcolor{red!20}63.6 \\
        \arrayrulecolor{lightgray}\hdashline
        NVILA-Lite-8B~\cite{liu2024nvila} & 43.2 & 41.4 & 42.8 \\
        NVILA-Lite-8B + \datasetName & 56.4 & \cellcolor{red!20}73.3 & \cellcolor{yellow!20}60.5 \\
        \arrayrulecolor{black} 
        \bottomrule
    \end{tabular}%
    }
    \vspace{-1em}
    \label{tab:vlm4d}
\end{table}

\begin{table*}[t]
    \centering
    \caption{\textbf{Ablation study on dataset composition.} We ablate the effect of adding tracking tasks to Std-\datasetName by evaluating the resulting models on both our \datasetName-Bench and the external VLM4D dataset~\cite{zhou2025vlm4d}. See Section~\ref{subsec:datasets} for description of each question type.}
    \label{tab:feature_ablation}
    \resizebox{0.95\linewidth}{!}{
    \begin{tabular}{@{}cl *{15}{w{c}{2.2em}} | *{3}{w{c}{2.2em}}@{}}
    \toprule
    & & \multicolumn{3}{c}{\textbf{Camera Motion}} & \multicolumn{6}{c}{\textbf{Object Motion}} & \multicolumn{5}{c}{\textbf{3D Spatial Understanding}} & & \multicolumn{3}{c}{\textbf{VLM4D}} \\
    \cmidrule(lr){3-5} \cmidrule(lr){6-11} \cmidrule(lr){12-16} \cmidrule(lr){18-20}
    & \textbf{Method} & 
    \rotatebox{90}{(1) Cam.} & \rotatebox{90}{(2) Cam.-Obj. Chg.} & \rotatebox{90}{\textbf{Average}} &
    \rotatebox{90}{(3)Rotation} & \rotatebox{90}{(4) Direction} & \rotatebox{90}{(5) Agent} & \rotatebox{90}{(6) Moved Dist.} & \rotatebox{90}{(7) Obj.-Obj. Chg.} & \rotatebox{90}{\textbf{Average}} &
    \rotatebox{90}{(8) Depth} & \rotatebox{90}{(9) Obj.-Obj. Dist.} & \rotatebox{90}{(10) MV Depth} & \rotatebox{90}{(11) MV Obj. Dist.} & \rotatebox{90}{\textbf{Average}} &
    \rotatebox{90}{\textbf{Overall Avg.}} &
    \rotatebox{90}{Real} & \rotatebox{90}{Synthetic} & \rotatebox{90}{\textbf{Average}} \\
    
    \midrule

    \textbf{(I)} & Qwen2.5-VL-3B~\cite{bai2025qwen2}& 50.4 & 41.5 & 47.1 & 33.6 & 39.2 & 27.3 & 40.0 & 29.2 & 36.9 & 58.5 & 56.3 & 49.3 & 47.0 & 54.4 & 46.7 & 48.2 & 35.3 & 45.0 \\
    \textbf{(II)}& \textbf{(I)} + Std-\datasetName & 79.6 & \cellcolor{red!20}86.3 & 82.0 & \cellcolor{red!20}76.1 & \cellcolor{red!20}78.0 & 81.8 & 77.1 & \cellcolor{red!20}80.8 & \cellcolor{red!20}78.2 & \cellcolor{red!20}89.7 & 84.2 & 92.6 & \cellcolor{red!20}83.6 & \cellcolor{red!20}88.0 & \cellcolor{red!20}83.1 & 54.4 & 47.6 & 52.7 \\
    \textbf{(III)}& \textbf{(II)} + True Motion Track (TM)& 81.0 & 81.5 & 81.3 & 72.6 & 71.6 & \cellcolor{red!20}90.9 & 78.6 & 73.1 & 73.9 & 88.8 & 85.2 & 91.9 & 78.4 & 86.8 & 81.3 & \cellcolor{red!20}55.0 & \cellcolor{red!20}56.9 & \cellcolor{red!20}55.5 \\
    \textbf{(IV)}& \textbf{(II)} +  Point Track (PT)& 81.0 & 82.7 & 82.3 & 72.6 & 70.3 & 72.7 & \cellcolor{red!20}79.3 & 80.0 & 73.9 & 88.5 & \cellcolor{red!20}85.8 & \cellcolor{red!20}93.2 & 79.9 & 87.3 & 82.0 & 51.7 & \cellcolor{red!20}56.9 & 53.0 \\
    \textbf{(V)}& \textbf{(II)} +  PT + TM& \cellcolor{red!20}83.0 & 84.5 & \cellcolor{red!20}82.8 & 71.7 & 73.3 & 72.7 & \cellcolor{red!20}79.3 & \cellcolor{red!20}80.8 & 75.4 & 87.9 & \cellcolor{red!20}85.8 & 92.6 & 79.9 & 87.0 & 82.4 & 52.3 & 51.9 & 52.2 \\
    \midrule
    
    \textbf{(I)} & Qwen2.5-VL-7B~\cite{bai2025qwen2}& 45.0 & 30.7 & 39.5 & 54.9 & 43.5 & 27.3 & 40.0 & 36.9 & 45.1 & 60.3 & 59.6 & 48.0 & 50.0 & 56.1 & 46.6 & 52.9 & 50.6 & 52.3 \\
    \textbf{(II)}& \textbf{(I)} + Std-\datasetName & 83.0 & \cellcolor{red!20}86.9 & 84.6 & 74.3 & \cellcolor{red!20}85.8 & 72.7 & 80.0 & \cellcolor{red!20}85.4 & \cellcolor{red!20}82.3 & 88.2 & 84.2 & 93.2 & \cellcolor{red!20}85.8 & 87.8 & \cellcolor{red!20}85.1 & 60.5 & 66.3 & 61.9 \\
    \textbf{(III)}& \textbf{(II)} + True Motion Track (TM)& 83.7 & 85.7 & 84.4 & \cellcolor{red!20}75.2 & 81.0 & 72.7 & 79.3 & 79.2 & 79.6 & 90.0 & 84.7 & 91.9 & 83.6 & 88.1 & 84.3 & \cellcolor{red!20}60.6 & 73.0 & \cellcolor{red!20}63.6 \\
    \textbf{(IV)}& \textbf{(II)} +  Point Track (PT)& 83.7 & 83.0 & 82.9 & \cellcolor{red!20}75.2 & 81.0 & \cellcolor{red!20}81.8 & \cellcolor{red!20}80.7 & 76.9 & 79.4 & 89.4 & \cellcolor{red!20}86.3 & \cellcolor{red!20}93.9 & 84.3 & \cellcolor{red!20}88.7 & 83.9 & 59.6 & 69.2 & 62.0 \\
    \textbf{(V)}& \textbf{(II)} +  PT + TM& \cellcolor{red!20}85.4 & 85.7 & \cellcolor{red!20}85.5 & 72.6 & 81.0 & 63.6 & 74.3 & 83.1 & 78.7 & \cellcolor{red!20}91.8 & 83.1 & 92.6 & 82.1 & 88.3 & 84.5 & 57.8 & \cellcolor{red!20}73.7 & 61.7 \\
    
    \midrule
    \textbf{(I)} & NVILA-Lite-8B~\cite{liu2024nvila}& 50.1 & 34.0 & 42.4 & 16.8 & 30.2 & 18.2 & 28.6 & 20.8 & 26.0 & 55.6 & 54.1 & 58.1 & 53.7 & 55.4 & 42.3 & 43.2 & 41.4 & 42.8 \\
    \textbf{(II)}& \textbf{(I)} + Std-\datasetName & 84.2 & 85.1 & 84.0 & 77.9 & \cellcolor{red!20}89.2 & 72.7 & 77.9 & 83.1 & 84.1 & \cellcolor{red!20}92.4 & 86.9 & 93.9 & 83.6 & \cellcolor{red!20}89.9 & 85.9 & 54.9 & 56.4 & 55.3 \\
    \textbf{(III)}& \textbf{(II)} + True Motion Track (TM)& 82.0 & 85.1 & 83.5 & 76.1 & 86.6 & 63.6 & 76.4 & 76.9 & 81.6 & 90.3 & 86.3 & 93.9 & 81.3 & 88.6 & 84.4 & \cellcolor{red!20}56.4 & 73.3 & \cellcolor{red!20}60.5 \\
    \textbf{(IV)}& \textbf{(II)} +  Point Track (PT)& 80.3 & 85.4 & 82.8 & \cellcolor{red!20}79.6 & 84.5 & 63.6 & 75.0 & 80.0 & 81.4 & 91.5 & 84.7 & \cellcolor{red!20}94.6 & \cellcolor{red!20}85.8 & 89.6 & 84.5 & 54.4 & 63.6 & 56.7 \\
    \textbf{(V)}& \textbf{(II)} +  PT + TM& 82.7 & 88.1 & 85.1 & 73.5 & 88.8 & 54.5 & 77.1 & 81.5 & 82.4 & 91.2 & 85.2 & 93.2 & 82.8 & 88.8 & 85.4 & 55.4 & 66.3 & 58.1 \\
    \textbf{(VI)}& \textbf{(III)} + Geometry Enc.~\cite{badki2025l4p}& \cellcolor{red!20}88.3 & \cellcolor{red!20}91.3 & \cellcolor{red!20}87.9 & 78.8 & 88.8 & \cellcolor{red!20}81.8 & \cellcolor{red!20}87.9 & \cellcolor{red!20}88.5 & \cellcolor{red!20}86.5 & 90.9 & \cellcolor{red!20}86.9 & 91.9 & 78.4 & 88.1 & \cellcolor{red!20}87.7 & 51.8 & \cellcolor{red!20}85.4 & 60.0 \\
    
    \bottomrule
    \end{tabular}
    }
\end{table*}

\vspace{-1em}\paragraph{Performance of off-the-shelf VLMs on our benchmark.}
Open-source VLMs that are strong on standard semantic benchmarks on average perform only slightly better than random guessing on \datasetName-Bench: most 7B--8B models lie in the low-to-mid $40\%$ range, compared to a $40.8\%$ random baseline.
Interestingly, in several categories, the baseline models are consistently biased toward wrong answers across samples, resulting in a performance worse than the random baseline.
For example, Qwen2.5-VL-7B attains only $30.7\%$ on camera-object questions and $27.3\%$ on agent-motion questions, and NVILA-Lite-8B achieves $16.8\%$ on object-rotation and $20.8\%$ on object-object distance, all far below the random baselines for the corresponding categories.
This behavior is also evident qualitatively: Figure~\ref{fig:teaser} shows cases where the baseline model systematically mispredicted simple attributes such as whether the camera is moving or whether the distance to an object is increasing or decreasing.
These results indicate that current open-source VLMs have limited capabilities for understanding 3D structure or object dynamics in the environment. %
On the other hand, some proprietary models perform noticeably better: Gemini-2.5-Pro reaches $66.8\%$ overall, substantially outperforming all open-source baselines.

\vspace{-1em}\paragraph{Training on \datasetName yields consistent gains.}
Fine-tuning on \datasetName dramatically improves performance for multiple VLM baselines.
Qwen2.5-VL-3B improves from $46.7\%$ to $81.3\%$ overall ($+34.6$ points), Qwen2.5-VL-7B from $46.6\%$ to $84.3\%$ ($+37.7$), and NVILA-Lite-8B from $42.3\%$ to $84.4\%$ ($+42.1$).
The gains are consistent across all categories: camera-motion averages rise from the low $40\%$s to above $80\%$, object-motion averages from $20$--$40\%$s to $70$--$80\%$s, and 3D spatial understanding averages from $\approx 55\%$ to above $86\%$.
Among all evaluated models, NVILA-Lite-8B+\datasetName achieves the highest overall score ($84.4\%$).
All three fine-tuned models perform better than the strongest proprietary baseline (Gemini-2.5-Pro) by about $14$--$18$ points in overall accuracy.
These results show that \datasetName is effective at teaching standard VLM architectures to perform fine-grained 4D perception.

\subsection{Results on VLM4D Benchmark}
To assess how well training on our dataset generalizes, we further evaluate on VLM4D~\cite{zhou2025vlm4d}, a 4D reasoning benchmark.
The benchmark consists of two splits: a real split, containing egocentric~\cite{grauman2022ego4d}, YouTube~\cite{xu2018youtube}, and DAVIS~\cite{pont2017davis} videos; and a synthetic split with samples generated by a video generative model~\cite{agarwal2025cosmos}.
VLM4D covers QA pairs that emphasize translational and rotational motion, perspective awareness, and motion continuity.

We train NVILA-Lite-8B, Qwen2.5-VL-7B, and Qwen2.5-VL-3B on our dataset and evaluate their performance on VLM4D, along with the other models from our benchmark.
All models are evaluated using exact string matching, following~\cite{yang2025thinking}, to handle minor answer variations, and final scores are reported as accuracy.

Table~\ref{tab:vlm4d} shows that before training on our dataset, Gemini-2.5-Pro leads with 62.8\% accuracy overall, while Qwen3-VL-32B is the best open-source model at $56.8\%$.
Fine-tuning NVILA-Lite-8B, Qwen2.5-VL-7B, and Qwen2.5-VL-3B on our dataset yields gains on both real and synthetic splits: NVILA-Lite-8B jumps from $42.8\%$ to $60.5\%$, Qwen2.5-VL-7B from $52.3\%$ to $63.6\%$, and Qwen2.5-VL-3B from $45.0\%$ to $55.5\%$.
After fine-tuning, Qwen2.5-VL-7B outperforms all models overall.
Notably, the smaller Qwen2.5-VL-3B model now surpasses the off-the-shelf Qwen2.5-VL-7B and achieves performance on par with the larger 32B models Qwen3-VL-32B and Qwen2.5-VL-32B.
These results highlight the effectiveness of our dataset for improving 4D understanding. %
We show additional benchmark results in the supplementary.

\subsection{Ablation Study}
We conduct ablations to understand the impact of tracking tasks and geometry features on VLM performance.

\vspace{-1em}\paragraph{True-motion tracking helps 4D reasoning.}
To better understand the impact of visual point tracking and our newly introduced true-motion point tracking tasks on VLM performance, we conduct a series of ablation studies exploring four distinct training settings.
Specifically, we begin with a baseline model trained solely on standard QA questions, which cover camera motion, object motion, and 3D spatial reasoning (Std-\datasetName in Table~\ref{tab:feature_ablation}). 
We then systematically augment this dataset by: (1) adding visual point tracking (PT) questions only, (2) adding true-motion point tracking (TM) questions only, and (3) adding both PT and TM questions to the training mix.
For each of these four settings, models are trained using the same procedures described in Section~\ref{subsec:training_details}, and the overall number of training samples remains constant.
Specifically, when introducing tracking questions, we replace $20\%$ of the standard QA samples with tracking-related questions; if both PT and TM questions are included, each type constitutes $10\%$ of the dataset. 
We set this to $20\%$, as higher values hurt standard QA performance.
We train NVILA-Lite-8B, Qwen2.5-VL-7B, and Qwen2.5-VL-3B on each of these four dataset combinations and evaluate their performance on both our benchmark (\datasetName-Bench) and VLM4D.
Overall, our experiments show that adding tracking tasks enables models to maintain almost all of their performance on standard QA questions in our benchmark, with only a slight decrease, while improving results on the VLM4D benchmark.
On average across all baselines, the best results on VLM4D are achieved when we use only the true-motion point tracking task (Row \textbf{III} for each baseline).
Given that VLM4D is specifically designed to benchmark 4D understanding under camera movement, the observed boost in performance from training on the true-motion tracking task validates the crucial role of this newly introduced task for 4D understanding in VLMs.

\vspace{-1em}\paragraph{Using geometry features improves performance.}
For the NVILA-Lite-8B model, we additionally train a geometry-augmented variant by adding L4P~\cite{badki2025l4p} geometry features to the model, using the same dataset mixture as the model with true-motion point tracking (\textbf{(III)}).
We observe a noticeable improvement on \datasetName-Bench and the VLM4D synthetic split, achieving $85.4\%$ on the VLM4D synthetic split. This highlights the importance of geometry features for 4D understanding in VLMs.
However, results on the VLM4D real split show a degradation in performance.
We hypothesize that this performance drop is due to the length of the real videos in VLM4D; using a uniform 32-frame sampling does not provide a high enough frame rate for the L4P encoder, which was specifically trained on densely sampled continuous videos.

\begin{figure}[t]
    \centering
    \includegraphics[width=0.95\columnwidth]{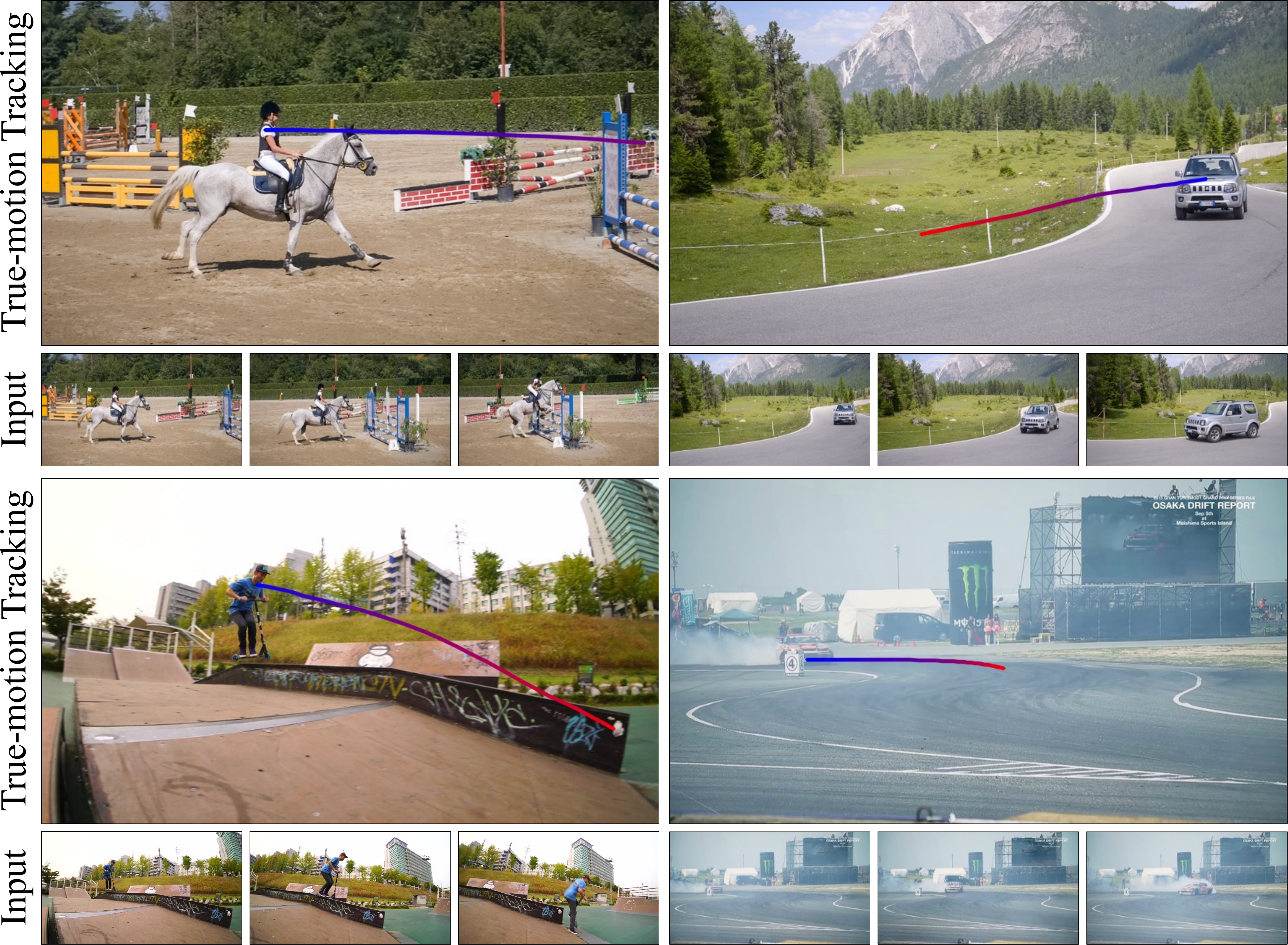}
    \caption{\textbf{Visualization of true motion track prediction.} True-motion tracking for dynamic scenes with camera motion (all but bottom right). Estimated tracks disentangle camera motion and summarize object motion as seen in the first frame.}
    \label{fig:truemotion_result}
    \vspace{-1em}
\end{figure}

\subsection{True-Motion Tracking Visualization}
Figure~\ref{fig:truemotion_result} presents qualitative results of true-motion tracking in challenging real-world scenes from~\cite{pont2017davis}.
We show outputs from the NVILA-Lite-8B model trained on our dataset, with the prompt: \textit{``Project the 3D trajectory of \{object name\} onto the image plane of frame 0. \ldots"}.
In each case, the estimated true-motion tracks successfully capture object motion relative to the first frame, providing an intuitive output that effectively summarizes dynamic object behavior despite the presence of camera movement.

\section{Conclusion}

We present a comprehensive framework to equip VLMs with better 4D understanding.
We implement a scalable QA generation pipeline, and collect data from a variety of sources to construct \datasetName, a large-scale 4D understanding dataset. 
We also introduce a new visual perception task, true-motion point tracking, to further encourage fine-grained 4D understanding in the presence of entangled object and camera motion. 
We demonstrate performance gains from fine-tuning standard VLM architectures on our dataset, and show improved generalization on VLM4D, an external 4D reasoning benchmark. 
Our ablation studies confirm the effectiveness of including true-motion tracking and further demonstrate that integrating geometric features into the VLM leads to additional improvements.

\section*{Acknowledgments}
We would like to thank Ligeng Zhu and Guanqi Zhan for discussions about training for the tracking task, and Zhiding Yu and Jan Kautz for their guidance.

\clearpage
\appendix
\twocolumn[{%
\centering
{\Large\textbf{4DP-QA: Scalable QA for 4D Perception in Vision Language Models}}\\[0.4em]
{\large\textbf{--- Supplementary Material ---}}
\vspace{1.5em}
}]
\vspace{-1em}\section{Additional Training Details}

\subsection{Architectural Details for the Model with L4P}

\begin{figure}[t]
    \centering
    \includegraphics[width=\columnwidth]{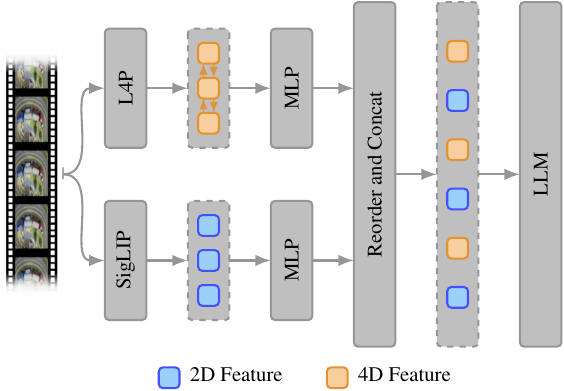}
    \caption{\textbf{Architecture of the VLM model enhanced with L4P.} We use the features extracted from L4P~\cite{badki2025l4p} as 4D geometry inputs to the MLLM~\cite{liu2024nvila}. The features from L4P are projected into the LLM input space and then used as input to the LLM along with visual tokens from SigLIP~\cite{zhai2023sigmoid}.}
    \label{fig:architecture}
\end{figure}

Beyond improving the baseline VLM models, we also investigate the effect of incorporating a geometry encoder (L4P~\cite{badki2025l4p}) into the VLM model.
Figure~\ref{fig:architecture} depicts the architecture of the VLM model augmented with L4P.
The features extracted from L4P serve as 4D geometric inputs to the VLM.
Specifically, we take the output of the VideoMAE backbone in L4P with shape $(T, H_{4D}, W_{4D}, C_{4D})$ and project it into the LLM input space using an MLP projector, following the design of the NVILA-Lite-8B projector~\cite{liu2024nvila}.
The projected features are then tokenized and flattened into a per-frame sequence of tokens with shape $(T, L_{4D}, C_{LLM})$, where $C_{LLM}$ is the LLM embedding dimension. 
For each frame $t$, we concatenate these L4P 4D tokens with the corresponding SigLIP~\cite{zhai2023sigmoid} visual tokens from NVILA along the token dimension, and finally build the full input sequence by stacking these per-frame token blocks in temporal order ($t{=}0,1,\ldots,T{-}1$). 

\subsection{Training L4P for True-Motion Point Tracking}
We describe how to train L4P to directly estimate true-motion point tracks without explicitly predicting depth or camera poses. The L4P model comprises a video encoder and task-specific decoders for both dense tasks (such as depth, optical flow, camera pose, and dynamic segmentation) and sparse tasks (including 2D and 3D visual point tracking). To enable 2D true-motion point tracking, we modify the sparse tracking head, which was originally designed for 2D visual point tracking.

The sparse tracking head in L4P uses the prompt encoding and mask decoding mechanism of Segment Anything~\cite{kirillov2023sam}. 
It introduces special input and output tokens that interact with the video encoder's output to estimate the 2D locations of points and their visibility.
For a given input query pixel $\vect{p}[t_q] = (t_q, x_q, y_q)$, the system embeds it using 3D positional encoding alongside a learnable embedding to produce an input point token.
In the case of 2D visual point tracking ($P_{2D}$ in Equation 1 in the main paper), L4P utilizes two output tokens with learnable embeddings: one for the 2D track position and another for track visibility.
These input and output tokens interact with each other and with the output of the video encoder through a two-way attention mechanism, allowing the model to decode the video features and generate the 2D visual point track $P_{2D}$.

To extend L4P for true-motion point tracking ($M_{2D}$ in Equation 2 in the main paper) for the input query pixel $\vect{p}[t_q]$, we simply introduce two additional output tokens with learnable embeddings--one for the position of the 2D true-motion point track, and one for its visibility.
We use the same two-way attention mechanism to simultaneously generate both the 2D visual point track and the 2D true-motion point track.
Ground-truth annotations for the true-motion point track are generated by projecting the GT 3D trajectory onto the camera view at time $t_q$.
The losses for the 2D true-motion point track mirror those used for the visual point track: an L1 loss for position and a binary cross-entropy loss for visibility.
Training follows the same data and procedures as the original L4P model.

\begin{figure}[t]
    \centering
    \includegraphics[width=\columnwidth]{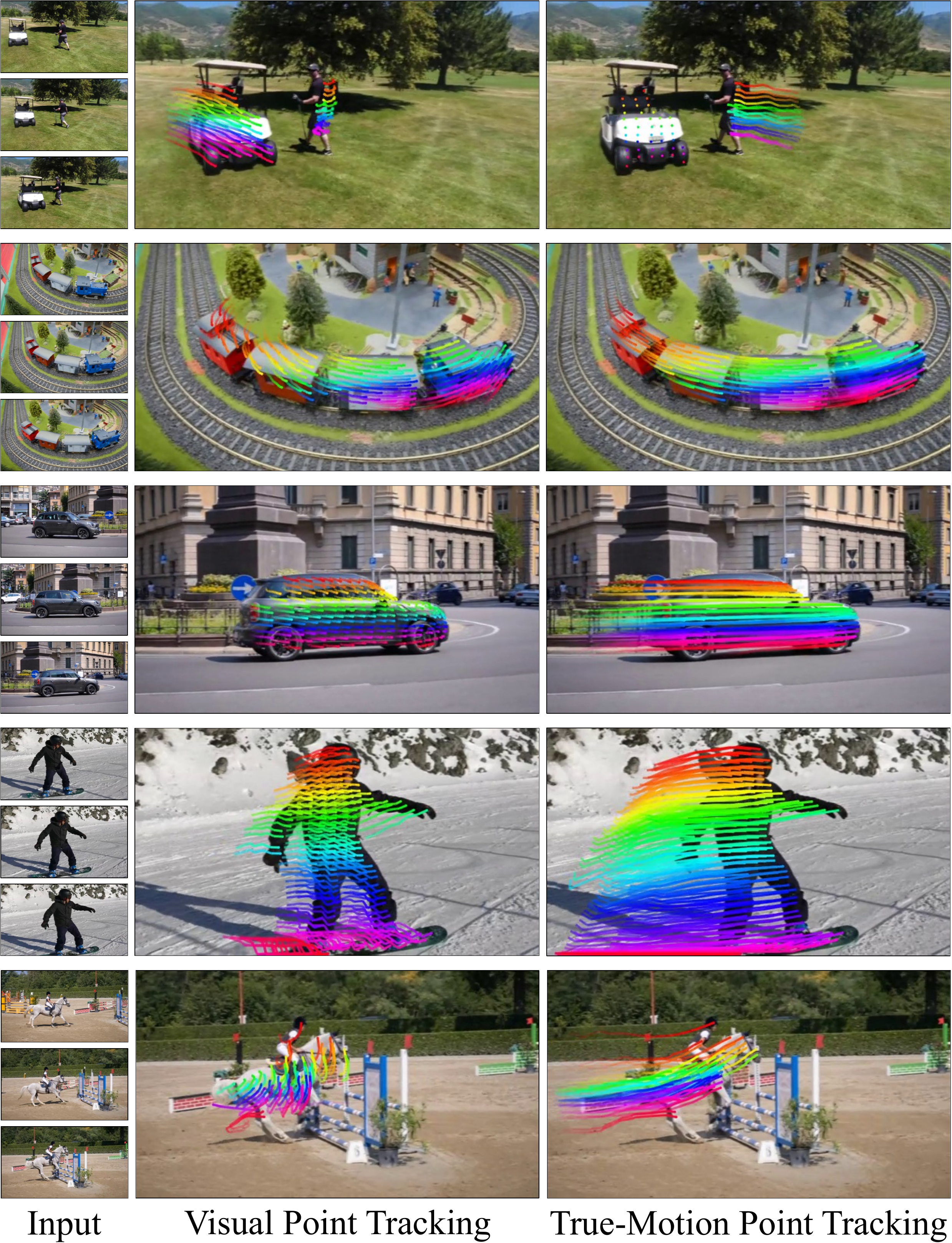}
    \caption{\textbf{Visual vs. True-motion Point Tracking using L4P.} 
    In the first row, true-motion point tracks correctly identify that the cart remains stationary while the person moves toward it.
    In the second row, true-motion point tracks for the train are better aligned with the railway track.
    Across other examples, true-motion point tracks provide a more accurate and interpretable representation of object movement as compared to the visual point tracks.
    }
    \label{fig:l4p_tracking}
\end{figure}

By adding true-motion to the L4P training and then incorporating it with the VLM, we saw an average increase from $85.3\%$ to $87.7\%$ on our benchmark and $59.5\%$ to $60.0\%$ on VLM4D.
Figure~\ref{fig:l4p_tracking} presents L4P results for both visual and true-motion point tracking on in-the-wild videos that involve complex camera and object movements. 
For each video, we select multiple 2D query points in the chosen frame and generate the corresponding visual and true-motion point tracks.
These results demonstrate that true-motion point tracking offers an intuitive and interpretable motion representation. 
Moreover, these results indicate that, although true-motion point tracking requires the model to implicitly reason about depth and camera poses, it is possible to learn this task directly--without explicit depth and pose computation or explicit geometric operations.
This suggests that VLM models may also be capable of learning such representations in a similar manner.

\section{\datasetName Details}
This section details the construction of \datasetName. We begin with our common design principles in Section~\ref{supp:sec:design_principles} and pipeline details in Section~\ref{supp:sec:pipeline_details}. Section~\ref{supp:sec:question_gen_pipeline} describes the question generation framework, including specific protocols for camera motion (Section~\ref{supp:sec:camera_motion_qas}), object motion (Section~\ref{supp:sec:object_motion_qas}), and 3D spatial understanding (Section~\ref{supp:sec:spatial_qas}). Point tracking questions are detailed in Section~\ref{supp:sec:point_tracking_qas}, followed by more details on QA generation in Section~\ref{supp:sec:answer_generation}.

\subsection{Common Design Principles}
\label{supp:sec:design_principles}

\paragraph{The use of hysteresis thresholding.}
To generate reliable ground truth annotations, we employ hysteresis thresholding with two threshold levels for each motion type. For translational motion, we define an upper threshold $\tau_{\text{trans}}^{\text{upper}}$ and a lower threshold $\tau_{\text{trans}}^{\text{lower}}$. A motion is classified as positive if the displacement exceeds $\tau_{\text{trans}}^{\text{upper}}$, negative if it falls below $\tau_{\text{trans}}^{\text{lower}}$, and undecided otherwise. Similarly, for rotational motion we use angular thresholds $\tau_{\text{rot}}^{\text{upper}}$ and $\tau_{\text{rot}}^{\text{lower}}$. Importantly, we designed all question-answer pairs to exclude undecided cases. Specifically, no questions ask about motions that fall in the undecided range (i.e., between the lower and upper thresholds), and undecided motion states do not appear as distractors in multiple-choice questions. This approach ensures high annotation quality by excluding ambiguous cases. 

\paragraph{Interval monotonicity constraint.}
For all motion-related questions, we enforce a monotonicity constraint to ensure that trajectories exhibit consistent directional change. Given a trajectory $\{\xi_t\}_{t=1}^T$ (representing translation along an axis or rotation angle), we require that all intermediate values lie between the endpoint values. Formally, let $\xi_{\min} = \min_t \xi_t$ and $\xi_{\max} = \max_t \xi_t$ denote the minimum and maximum values. The trajectory is considered monotonic if either $\xi_1 \approx \xi_{\min}$ and $\xi_T \approx \xi_{\max}$, or $\xi_1 \approx \xi_{\max}$ and $\xi_T \approx \xi_{\min}$.

To account for noise and minor fluctuations, we introduce a tolerance margin $\delta = \gamma (\xi_{\max} - \xi_{\min})$, where $\gamma = 0.1$ is the margin ratio. The approximation $\xi_i \approx \xi_j$ is satisfied if $|\xi_i - \xi_j| \leq \delta$. This constraint filters out trajectories with severe non-monotonic behavior, ensuring that the ground truth annotations correspond to clear, unambiguous motion patterns. The monotonicity check is applied to all directional translation components, rotation angles, and distance trajectories before generating questions.

\paragraph{Balanced sampling of static and dynamic objects.}
To avoid dataset bias toward moving objects, we implement balanced sampling when selecting object pairs. For questions involving two objects, we preferentially sample pairs consisting of one moving and one static object. This ensures that models learn to reason about both dynamic and static elements in the scene.

\paragraph{Gravity-aligned motion sensing.}
To ensure consistent geometric reasoning across datasets with different coordinate conventions, we identify the gravity axis for each dataset and align motion analysis to this axis for camera and object motion questions. We decompose translations and rotations into gravity-aligned components. Let $\mathbf{g} \in \mathbb{R}^3$ denote the gravity direction vector (pointing upward). For any 3D motion vector $\mathbf{v}$, we decompose it into vertical and horizontal components: $\mathbf{v}_\text{vert} = (\mathbf{v} \cdot \mathbf{g})\mathbf{g}$ and $\mathbf{v}_\text{horiz} = \mathbf{v} - \mathbf{v}_\text{vert}$. This enables intuitive motion descriptions such as ``moving forward" and ``moving upward" that are invariant to camera orientation.

\subsection{Pipeline Details}
\label{supp:sec:pipeline_details}

\subsubsection{Data Standardization}
\begin{table}[h]
\centering
\resizebox{\linewidth}{!}{%
\begin{tabular}{l p{10cm}}
\toprule
\textbf{Component} & \textbf{Description} \\
\midrule
Video Frames & RGB video frames with shape $(T, H, W, 3)$ where $T$ is the number of frames. \\
Frame Rate & Frames per second of the video sequence. \\
Camera Intrinsics & Camera intrinsic matrix $\mat{K} \in \mathbb{R}^{3 \times 3}$ (assumed shared across frames), encoding focal lengths and principal point. \\
Camera Extrinsics & Camera extrinsic matrices $\mat{T}[t] \in \mathbb{R}^{4 \times 4}$ for each frame, representing world-to-camera transformation. \\
Depth Maps & Per-pixel depth measurements $\mathbf{D}[t] \in \mathbb{R}^{H \times W}$ indicating distance from the camera. Depth values are not necessarily valid for all pixels; the maps may be sparse. \\
Depth Validity Masks & Binary masks $\mathbf{B}[t] \in \{0,1\}^{H \times W}$ indicating reliable depth values, where 1 indicates a valid depth value and 0 indicates an invalid depth value. \\
Instance Segmentation & Per-pixel segmentation masks $\mathbf{S}[t] \in \mathbb{Z}^{H \times W}$ assigning each pixel to an object instance \\
Instance Metadata & Semantic information for each tracked object, including its name and category. If the category is missing or incomplete in the dataset, it can be generated using an off-the-shelf VLM~\cite{yang2025qwen3}. \\
Object Poses & Object-to-world transformation matrices $\mat{O}_{i}[t] \in \mathbb{R}^{4 \times 4}$ encoding rigid-body pose for each tracked object $i$ in frame $t$. \\
\bottomrule
\end{tabular}%
}
\caption{\textbf{Standardized data format specification.} 
We unify the formats of datasets from various sources into a single standardized format to streamline the pipeline.
}
\label{tab:unified_format}
\end{table}
We standardize all datasets from various sources into a unified format to streamline the generation pipeline, as summarized in Table~\ref{tab:unified_format}. Particular care is taken to unify the coordinate conventions, since different datasets adopt different standards. We adopt the OpenCV coordinate convention: in image space, the x-axis points to the right and the y-axis points downward, while in the camera coordinate system, the z-axis points forward. For depth, we convert all data to planar depth when a dataset provides only radial depth.
We also observe that each dataset aligns gravity with one of the coordinate axes, but the specific axis differs across datasets. We identify the gravity axis for each dataset and incorporate this information during dataset generation to maintain consistent geometric reasoning.

Some datasets contain incomplete annotations for certain data types we require. For example, HOT3D~\cite{banerjee2024introducing} provides depth only for the hands and objects, leaving other regions invalid. In such cases, we avoid generating questions that reference regions where depth is unavailable. 

Several datasets also provide incomplete or missing instance names. To address this, we use the Qwen3-VL-32B-Instruct~\cite{yang2025qwen3} model to annotate object descriptions. Specifically, for each object, we draw a red bounding box on the frame where the entire object is visible and appears largest in the image, and we prompt the VLM to identify the object's name. The exact prompt used for this annotation process is as follows: 
\begin{tcolorbox}[colback=gray!5,colframe=gray!40,title=Prompt for Object Name Annotation]
\small
\textit{%
Give a short description of the object inside the red bounding box. Keep it concise and to the point. If applicable specify the object category, its attributes like colors. Describe in a way that is easy to distinguish from other objects in the image. Do not include the description of the surrounding of the object. Use less than 8 words.%
}
\end{tcolorbox}
In cases where the dataset contains noisy or unreliable metadata for certain objects, we append an additional prompt to guide the model. This prompt provides the existing metadata as a hint but instructs the model to use it only if it is helpful for producing an accurate visual description.
\begin{tcolorbox}[colback=gray!5,colframe=gray!40,title=Prompt for Handling Noisy Metadata]
\small
\textit{
I also have some metadata that may or may not be correct. For the object inside the red box the current metadata is \{metadata\}. Only use this metadata if it helps with visual description, otherwise ignore it.
}
\end{tcolorbox}

\subsubsection{Asset Sampling} 
We employ a two-stage approach for sampling valid question-answer pairs. 

\vspace{-1em}\paragraph{Stage 1: Interesting segment identification.} We identify \textit{interesting segments} where significant motion occurs to focus on informative video portions. First, we calculate the maximum spatial occupancy of each object across the video using pooled segmentation masks. Objects exceeding a specific area ratio threshold (default 0.1\%) are classified as \textit{large objects}. We then filter this set to include only objects that are moving, based on their 3D displacement. A contiguous sequence of frames is marked as an \textit{interesting segment} if it contains at least one large, moving object in every frame for a minimum duration.

\vspace{-1em}\paragraph{Stage 2: Snippet sampling within segments.} For each sampled interesting segment, we select a sub-sequence (snippet) of frames. If the segment length is shorter than the target snippet length, we pad the selection by extending the start and end indices into the surrounding video to meet the required length. Otherwise, we sample a sub-segment with a randomized temporal span within the interesting segment boundaries. We then generate evenly spaced frame indices between the selected start and end frames. If the number of sampled frames exceeds the target input size (typically 32), we uniformly subsample the sequence. Finally, to increase data diversity, we apply temporal reversal augmentation with a probability of 0.5, flipping the chronological order of the frames.

\subsubsection{Object Reference System}

To refer to objects and points in video frames, we implement a flexible reference system that supports visual markers, coordinate-based specifications, and natural language names.
This system enables clear, unambiguous references while maintaining diversity in question phrasing.

\vspace{-1em}\paragraph{(1) Color-coded circular markers.} Objects are referenced using colored circles drawn around their segmentation masks. For a given object with track ID $i$ in frame $f_t$, we compute its bounding box from the segmentation mask $\mathbf{S}[f_t]$ and draw an ellipse with center $\mathbf{c}_{2\text{D}}$ at the box center and axes determined by the box dimensions. 
We follow the format of the circle described in~\cite{shtedritski2023does}.
The circle is drawn with thickness $\tau_{\text{thick}} = 0.01 \cdot H$ where $H$ is the image height. We use a fixed color palette that cycles through six colors: \{red, green, blue, yellow, purple, orange\}. Example references include ``the red circle in frame 2" or ``the object at the blue circle (frame 3)." 
For visual examples, please refer to the dataset visualizer described in Section~\ref{supp:sec:visualization}.

To clearly indicate an object using a circular marker, we select the most suitable frame within the snippet based on two criteria. First, we check which frames show the entire object inside the image, without any part cut off at the image edges. Second, among those frames, we pick the one where the object covers the largest area in its segmentation mask. If no frame shows the object fully inside the image, we instead choose the frame where the object's visible area is largest, even if it is partly outside the image. This approach ensures the object is both visible and clearly highlighted by the marker.

\vspace{-1em}\paragraph{(2) Point markers with color.} For point-based queries, following~\cite{shtedritski2023does}, we draw a circular marker at the specified pixel location $(x, y)$ with radius $r = 0.06 \cdot H$ and thickness $\tau_{\text{thick}}$, and add a filled dot of radius $\tau_{\text{thick}}$ at the center. The point is then referenced as ``the red point at (0.451, 0.328) in frame 0" or simply ``the green point (frame 1)."

\vspace{-1em}\paragraph{(3) Coordinate-based references.} 
Points can also be referenced purely by their normalized coordinates without visual markers: ``the point at (0.451, 0.328)" or ``point (0.451, 0.328) in frame 3." Coordinates are normalized to $[0, 1] \times [0, 1]$ where $(0, 0)$ is the top-left corner.
For our benchmark evaluation, however, we do not include questions that involve these absolute coordinates, as coordinate conventions (e.g., origin and normalization) may differ across different VLMs.

\vspace{-1em}\paragraph{(4) Natural language name.} Objects can be referenced directly by their semantic class name or description if available in the dataset metadata or generated by a VLM. Example references include ``the white sedan" or ``the person in the red shirt."

\subsection{Question Generation Pipeline}
\label{supp:sec:question_gen_pipeline}

The generation of each question-answer pair follows a structured pipeline that ensures diversity, balance, and annotation quality.

\vspace{-1em}\paragraph{Question type and snippet selection.} For each annotation, we first sample a question type from the available set (e.g., camera motion, object rotation, depth comparison) according to configured sampling weights. We then sample a snippet within a video $\mathcal{F} = \{f_1, \ldots, f_T\}$ as described in the asset sampling section.

\vspace{-1em}\paragraph{Template-based question generation.} Each question type maintains a set of question templates and choice templates. For example, the camera motion question type has templates such as ``Which of the following best describes the camera's movement?" and ``Considering the entire clip, which of the following describes the camera's primary motion?" During generation, we randomly select a template and populate it with the analysis results (e.g., detected motions, object references). This template-based approach ensures linguistic diversity while maintaining semantic consistency.

Below, we detail each question-generation pipeline, beginning with the definition of notations, followed by camera motion QAs, object motion QAs, and finally, 3D spatial understanding QAs.
\vspace{-1em}\paragraph{Notations.}
Throughout this section, we use the following notations aligned with Table~\ref{tab:unified_format}: $T$ denotes the number of frames in a snippet, indexed by $t \in \{1, \ldots, T\}$. Camera intrinsics are denoted $\mat{K} \in \mathbb{R}^{3 \times 3}$ (shared across frames), and camera extrinsics at frame $t$ are $\mat{T}[t] \in \mathbb{R}^{4 \times 4}$. The camera position in world coordinates is computed as $\mathbf{p}_{\text{cam}}(t) = -\mathbf{R}[t]^\top \mathbf{t}[t]$, where $\mat{T}[t] = [\mathbf{R}[t] | \mathbf{t}[t]]$. For object $i$, we denote its pose at frame $t$ as $\mat{O}_i[t] \in \mathbb{R}^{4 \times 4}$ and its center in world coordinates as $\mathbf{c}_i(t) = \mat{O}_i[t]_{:3, 3}$. The gravity direction is $\mathbf{g} \in \mathbb{R}^3$ with $\|\mathbf{g}\| = 1$.

\subsubsection{Camera Motion QAs}
\label{supp:sec:camera_motion_qas}

\paragraph{(1) Camera movement.}
We analyze camera motion by decomposing it into gravity-aligned translation and rotation components. For a snippet with frames $\mathcal{F} = \{f_1, \ldots, f_T\}$, we compute the camera trajectory $\{\mathbf{p}_{\text{cam}}(f_t)\}_{t=1}^T$ in world coordinates. We project this trajectory onto three axes: (1) the forward direction $\mathbf{d}_{\text{fwd}}$, obtained by projecting the initial camera's forward vector onto the ground plane; (2) the right direction $\mathbf{d}_{\text{right}} = \mathbf{d}_{\text{fwd}} \times \mathbf{g}$; and (3) the vertical direction $\mathbf{g}$. 
The camera translation is then determined from the signed displacements along these three projected trajectories.

For rotation analysis, we compute gravity-aligned Euler angles. Let $\mathbf{R}[t]$ denote the camera rotation matrix at frame $t$ (extracted from $\mat{T}[t]$). We define: (1) \textit{pitch} as the angle between the camera's forward vector and a plane perpendicular to the gravity axis, (2) \textit{yaw} as the rotation around the gravity axis, measured relative to an arbitrary horizontal direction that remains fixed throughout the snippet, and (3) \textit{roll} as the rotation around the camera's viewing axis. All angles are computed relative to the first frame. Motion is classified using hysteresis thresholding: $|\Delta| > \tau^{\text{upper}}$ indicates motion in one direction, $|\Delta| < \tau^{\text{lower}}$ indicates no motion, and intermediate values are discarded as undecided.

\begin{tcolorbox}[colback=blue!5,colframe=blue!40!black,title=\textbf{Example Question Template: Camera Movement}]
\small
\textbf{Question:} \textit{Which of the following best describes the camera's movement?}

\vspace{2mm}
\textbf{Choices:} 
\begin{itemize}[leftmargin=*, itemsep=1pt]
\item The camera is moving forward and panning right
\item The camera is moving backward and tilting down
\item The camera is stationary
\end{itemize}
\end{tcolorbox}

\paragraph{(2) Camera-object distance change.}
For camera-object proximity analysis, we track the distance between the camera and an object's center over time. Let $d_i(t) = \|\mathbf{c}_i(t) - \mathbf{p}_{\text{cam}}(t)\|$ denote the distance at frame $t$. We require the distance trajectory $\{d_i(f_t)\}_{t=1}^T$ to be monotonic. The change is classified based on $\Delta d = d_i(f_T) - d_i(f_1)$ and relative change $\rho = \Delta d / \max(d_i(f_1), d_i(f_T))$. The objects are getting closer if $\Delta d < -\tau_{\text{trans}}^{\text{upper}}$ and $\rho < -\tau_{\text{rel}}^{\text{upper}}$; moving apart if $\Delta d > \tau_{\text{trans}}^{\text{upper}}$ and $\rho > \tau_{\text{rel}}^{\text{upper}}$; and maintaining distance if $|\Delta d| < \tau_{\text{trans}}^{\text{lower}}$.

Additionally, we analyze the relative movement direction between camera and object on the ground plane. Let $\mathbf{v}_{\text{cam}} = \mathbf{p}_{\text{cam}}(f_T) - \mathbf{p}_{\text{cam}}(f_1)$ and $\mathbf{v}_{\text{obj}} = \mathbf{c}_i(f_T) - \mathbf{c}_i(f_1)$ be the displacement vectors, projected onto the ground plane. We compute the alignment $\alpha = \cos^{-1}(\hat{\mathbf{v}}_{\text{cam}} \cdot \hat{\mathbf{v}}_{\text{obj}})$. The directions are classified as: along (same direction) if $\alpha < \tau_{\text{align}}$, against (opposite) if $\alpha > 180° - \tau_{\text{align}}$, and perpendicular if $|\alpha - 90°| < \tau_{\text{perp}}$.

\begin{tcolorbox}[colback=blue!5,colframe=blue!40!black,title=\textbf{Example Question Template: Camera-Object Distance}]
\small
\textbf{Question:} \textit{Describe the change in proximity between the camera and the red cup (frame 5) throughout the clip.}

\vspace{2mm}
\textbf{Choices:} 
\begin{itemize}[leftmargin=*, itemsep=1pt]
\item The camera and the red cup got closer
\item The camera and the red cup moved further apart
\item The distance between them stayed roughly the same
\end{itemize}
\end{tcolorbox}

\subsubsection{Object Motion QAs}
\label{supp:sec:object_motion_qas}

\paragraph{(3) Rotation.}
We analyze object rotation from a top-down perspective by tracking yaw changes around the gravity axis. For object $i$, we extract rotation matrices $\{\mathbf{R}_i[f_t]\}_{t=1}^T$ from its pose $\mat{O}_i[f_t]$. To measure yaw, we project the object's local x-axis onto the ground plane at each frame, obtaining $\mathbf{x}_i^\perp(t) = \mathbf{x}_i(t) - (\mathbf{x}_i(t) \cdot \mathbf{g})\mathbf{g}$, where $\mathbf{x}_i(t) = \mathbf{R}_i[f_t] \mathbf{e}_1$. We measure angles against a fixed horizontal reference frame and unwrap the angle trajectory $\{\theta_i(f_t)\}_{t=1}^T$ to handle $\pm 180°$ discontinuities. The yaw change $\Delta\theta = \theta_i(f_T) - \theta_i(f_1)$ is classified as clockwise if $\Delta\theta < -\tau_{\text{rot}}^{\text{upper}}$, counter-clockwise if $\Delta\theta > \tau_{\text{rot}}^{\text{upper}}$, and no rotation if $|\Delta\theta| < \tau_{\text{rot}}^{\text{lower}}$. We verify that pitch and roll changes remain below $\tau_{\text{rot}}^{\text{upper}}$ to ensure valid top-down analysis.

\begin{tcolorbox}[colback=blue!5,colframe=blue!40!black,title=\textbf{Example Question Template: Object Rotation}]
\small
\textbf{Question:} \textit{From a top-down perspective, which of the following best describes the rotation of the blue toy car (frame 3)?}

\vspace{2mm}
\textbf{Choices:} 
\begin{itemize}[leftmargin=*, itemsep=1pt]
\item The blue toy car is rotating clockwise
\item The blue toy car is rotating counter-clockwise
\item The blue toy car is not rotating
\end{itemize}
\end{tcolorbox}

\paragraph{(4) Direction.}
For directional motion analysis, we decompose object translation in a stable reference frame defined by the initial camera view. Let $\mathbf{v}_i = \mathbf{c}_i(f_T) - \mathbf{c}_i(f_1)$ be the object's displacement vector. We project it onto three axes: (1) camera-aligned horizontal right $\mathbf{d}_{\text{right}}$, (2) gravity direction $\mathbf{g}$, and (3) camera-aligned horizontal forward $\mathbf{d}_{\text{fwd}}$. For each axis, we compute the projected trajectory and verify monotonicity. Motion is classified independently for each direction using thresholds that adapt to object depth: $\tau_i^{\text{upper}} = \max(\tau_{\text{trans}}^{\text{upper}} \cdot 2, \tau_{\text{depth\text{-}rel}}^{\text{upper}} \cdot d_i(f_1))$, where $d_i(f_1)$ is the initial depth. This accounts for the fact that objects farther from the camera require larger absolute displacements to be perceived as moving. Questions can ask about single-direction motion or combined motion (e.g., ``moving forward and to the right"). We also generate counting questions asking how many objects exhibit a specific directional motion.

\begin{tcolorbox}[colback=blue!5,colframe=blue!40!black,title=\textbf{Example Question Template: Directional Motion}]
\small
\textbf{Question:} \textit{From the camera's initial perspective (where `forward/backward' are relative to the camera's forward viewing direction), which option best describes the movement of the yellow block (frame 2)?}

\vspace{2mm}
\textbf{Choices:} 
\begin{itemize}[leftmargin=*, itemsep=1pt]
\item The yellow block is moving forward and up
\item The yellow block is moving backward and left
\item The yellow block is stationary
\end{itemize}
\end{tcolorbox}

\paragraph{(5) Agent motion.}
For agent motion (e.g., a person or car), we combine perspective-based translation with egocentric rotation analysis. First, we determine if the agent is moving forward or backward from its own perspective by comparing its velocity vector $\mathbf{v}_i$ with its forward direction $\mathbf{f}_i(f_1) = \mathbf{R}_i[f_1] \mathbf{e}_1$, both projected onto the ground plane. Let $\alpha = \cos^{-1}(\hat{\mathbf{v}}_i \cdot \hat{\mathbf{f}}_i)$ be the angle between them. The agent is moving forwards if $\alpha < \tau_{\text{persp}}^{\text{lower}}$, backwards if $\alpha > 180° - \tau_{\text{persp}}^{\text{lower}}$, and neither if $\tau_{\text{persp}}^{\text{lower}} < \alpha < \tau_{\text{persp}}^{\text{upper}}$ or $180° - \tau_{\text{persp}}^{\text{upper}} < \alpha < 180° - \tau_{\text{persp}}^{\text{lower}}$.

Second, we analyze turning direction using the yaw trajectory as described in rotation analysis. However, if the agent is moving backward, we flip the turn direction (left $\leftrightarrow$ right) to maintain consistency with the agent's egocentric perspective. Both translation and rotation trajectories must be monotonic for valid classification.

\begin{tcolorbox}[colback=blue!5,colframe=blue!40!black,title=\textbf{Example Question Template: Agent Motion}]
\small
\textbf{Question:} \textit{How's the agent's motion? Is it moving forwards, backwards, or turning left or right?}

\vspace{2mm}
\textbf{Choices:}
\begin{itemize}[leftmargin=*, itemsep=1pt]
\item The person is moving forwards and turning left
\item The person is moving backwards
\item The person is not moving forwards or backwards
\end{itemize}
\end{tcolorbox}

\paragraph{(6) Moved distance.}
To compare movement distances between objects, we compute the 3D displacement for each object: $\Delta_i = \|\mathbf{c}_i(f_T) - \mathbf{c}_i(f_1)\|$. For a pair of objects $(i, j)$, we classify their relative movement as: object $i$ moved further if $\Delta_i > \Delta_j + \tau_{\text{comp}}$; object $j$ moved further if $\Delta_j > \Delta_i + \tau_{\text{comp}}$; or neither moved significantly if $\Delta_i < \tau_{\text{trans}}^{\text{lower}}$ and $\Delta_j < \tau_{\text{trans}}^{\text{lower}}$. The comparison threshold $\tau_{\text{comp}} = 5 \cdot \tau_{\text{trans}}^{\text{upper}}$ ensures clear distinction between movements.

\begin{tcolorbox}[colback=blue!5,colframe=blue!40!black,title=\textbf{Example Question Template: Moved Distance}]
\small
\textbf{Question:} \textit{Which object moved a greater distance, the red ball (frame 1) or the green cube (frame 2)?}

\vspace{2mm}
\textbf{Choices:} 
\begin{itemize}[leftmargin=*, itemsep=1pt]
\item The red ball moved further
\item The green cube moved further
\item Neither object moved significantly
\end{itemize}
\end{tcolorbox}

\paragraph{(7) Object-object distance change.}
For analyzing distance changes between two objects, we compute the inter-object distance trajectory $\{d_{ij}(f_t)\}_{t=1}^T$, where $d_{ij}(t) = \|\mathbf{c}_i(t) - \mathbf{c}_j(t)\|$. We require valid 3D positions for both objects in at least 80\% of overlapping frames and monotonicity of the distance trajectory. The change is classified based on $\Delta d_{ij} = d_{ij}(f_T) - d_{ij}(f_1)$ with adaptive thresholding: $\tau_{ij}^{\text{upper}} = \max(\tau_{\text{trans}}^{\text{upper}} \cdot 2, \tau_{\text{depth\text{-}rel}}^{\text{upper}} \cdot \max(d_i(f_1), d_j(f_1)))$, where $d_i(f_1)$ and $d_j(f_1)$ are the distances of each object from the camera at the first frame. Objects are getting closer if $\Delta d_{ij} < -\tau_{ij}^{\text{upper}}$, moving apart if $\Delta d_{ij} > \tau_{ij}^{\text{upper}}$, and staying at the same distance if $|\Delta d_{ij}| < \tau_{\text{trans}}^{\text{lower}}$.

\begin{tcolorbox}[colback=blue!5,colframe=blue!40!black,title=\textbf{Example Question Template: Object-Object Distance}]
\small
\textbf{Question:} \textit{Which best describes the distance change between the white mug (frame 1) and the laptop (frame 3) over the course of the video?}

\vspace{2mm}
\textbf{Choices:} 
\begin{itemize}[leftmargin=*, itemsep=1pt]
\item They are getting closer
\item They are getting farther apart
\item They are staying at a similar distance
\end{itemize}
\end{tcolorbox}

\subsubsection{3D Spatial Understanding QAs}
\label{supp:sec:spatial_qas}

\paragraph{(8) Depth.}
For single-frame depth comparison, we select two points $\mathbf{p}_a$ and $\mathbf{p}_b$ in frame $f_t$ with valid depth values from $\mathbf{D}[f_t]$ and compute their distances to the camera: $d_a = \|\mathbf{p}_a - \mathbf{p}_{\text{cam}}(f_t)\|$ and $d_b = \|\mathbf{p}_b - \mathbf{p}_{\text{cam}}(f_t)\|$. The question asks which point (or object) is closer/farther to the camera. For objects, we use their center positions $\mathbf{c}_i(f_t)$ and $\mathbf{c}_j(f_t)$. Points are selected to ensure meaningful depth difference: $\max(d_a/d_b, d_b/d_a) > \tau_{\text{depth\text{-}ratio}}$, where $\tau_{\text{depth\text{-}ratio}} = 1.25$.

\begin{tcolorbox}[colback=blue!5,colframe=blue!40!black,title=\textbf{Example Question Template: Depth Comparison}]
\small
\textbf{Question:} \textit{In frame 0, which is closer to the camera: the red point (A) or the blue point (B)?}

\vspace{2mm}
\textbf{Choices:} A, B
\end{tcolorbox}

\paragraph{(9) Object-object distance.}
For single-frame object-object distance comparison, we select three objects $i$, $j$, and $k$ visible in frame $f_t$. We compute distances $d_{ij} = \|\mathbf{c}_i(f_t) - \mathbf{c}_j(f_t)\|$ and $d_{ik} = \|\mathbf{c}_i(f_t) - \mathbf{c}_k(f_t)\|$, and ask which of objects $j$ or $k$ is closer to object $i$. We ensure meaningful distance difference with the same ratio threshold.

\begin{tcolorbox}[colback=blue!5,colframe=blue!40!black,title=\textbf{Example Question Template: Object-Object Distance}]
\small
\textbf{Question:} \textit{In frame 0, which point is closer to the green cup: the yellow point (A) or the purple point (B)?}

\vspace{2mm}
\textbf{Choices:} A, B
\end{tcolorbox}

\paragraph{(10) Multi-view depth.}
For cross-frame depth comparison, we select two objects $i$ and $j$ that appear in different frames $f_a$ and $f_b$ respectively, and compare their depths relative to a reference frame $f_r$. Let $\mathbf{p}_{\text{cam}}(f_r)$ be the reference camera position. We compute $d_i = \|\mathbf{c}_i(f_r) - \mathbf{p}_{\text{cam}}(f_r)\|$ and $d_j = \|\mathbf{c}_j(f_r) - \mathbf{p}_{\text{cam}}(f_r)\|$. The objects are annotated with their best visible frames $f_a$ and $f_b$ (where they appear largest), but the depth comparison is performed at the shared reference frame $f_r$. This tests the model's ability to reason about 3D positions across different viewpoints.

\begin{tcolorbox}[colback=blue!5,colframe=blue!40!black,title=\textbf{Example Question Template: Multi-View Depth}]
\small
\textbf{Question:} \textit{In frame 2, which is closer to the camera: the orange bottle (A) shown in frame 0 or the blue pen (B) shown in frame 5?}

\vspace{2mm}
\textbf{Choices:} A, B
\end{tcolorbox}

\paragraph{(11) Multi-view object-object distance.}
For cross-frame object-to-object distance comparison, we select three objects $i$, $j$, $k$ where object $i$ serves as reference and objects $j$ and $k$ are presented in their best visible frames. All distance measurements are computed at a common reference frame $f_r$: $d_{ij} = \|\mathbf{c}_i(f_r) - \mathbf{c}_j(f_r)\|$ and $d_{ik} = \|\mathbf{c}_i(f_r) - \mathbf{c}_k(f_r)\|$. The question asks which of objects $j$ or $k$ is closer to reference object $i$ in frame $f_r$, even though the objects are shown in different frames.

\begin{tcolorbox}[colback=blue!5,colframe=blue!40!black,title=\textbf{Example Question Template: Multi-View Object Distance}]
\small
\textbf{Question:} \textit{In frame 1, which is closer to the white keyboard: the mouse (A) shown in frame 0 or the phone (B) shown in frame 3?}

\vspace{2mm}
\textbf{Choices:} A, B
\end{tcolorbox}

\subsubsection{Point Tracking QAs}
\label{supp:sec:point_tracking_qas}

Point tracking questions require generating temporally consistent 3D trajectories and their 2D projections for query points across frames. We first recover a unified 3D trajectory $\{\vect{X}[f_t]\}$ for each query point using our standardized 3D scene representation, and then obtain different 2D track representations by projecting this trajectory into appropriate camera views.

Given a query point $\vect{p}[f_q]$ at reference frame $f_q$, we generate its trajectory across all frames $f_t$. The algorithm consists of three main steps:

\vspace{-1em}\paragraph{Step 1: 3D Initialization.} We first unproject the query point to 3D world coordinates. Let $z_q = \mathbf{D}[f_q](\vect{p}[f_q])$ be the depth at the query point. We compute the 3D point in the camera coordinate system at frame $f_q$:
\begin{equation}
    \vect{X}_{\text{cam}}[f_q] = z_q \mat{K}^{-1} [\vect{p}[f_q]; 1]
\end{equation}
We then transform to world coordinates using the camera extrinsics: $\vect{X}[f_q] = \mat{T}[f_q]^{-1} [\vect{X}_{\text{cam}}[f_q]; 1]$. This gives us the 3D position of the query point in the world coordinate system at the query time.

\vspace{-1em}\paragraph{Step 2: Motion Association.} We determine whether the point belongs to a moving object or the static scene by querying the segmentation mask. Let $i = \mathbf{S}[f_q](\vect{p}[f_q])$ be the instance ID at the query point. If the point belongs to the background, it is part of the static environment and maintains the same world coordinates $\vect{X}[f_t] = \vect{X}[f_q]$ across all frames. Conversely, if the point belongs to a moving object, we transform it to the object's local coordinate system at frame $f_q$: $\vect{X}_{\text{local}} = \mat{O}_i[f_q]^{-1} [\vect{X}[f_q]; 1]$. This local representation is invariant to the object's motion. For each frame $f_t$, we then transform back to world coordinates using that frame's object pose: $\vect{X}[f_t] = \mat{O}_i[f_t] \vect{X}_{\text{local}}$, which correctly handles object motion, rotation, and deformation.

\vspace{-1em}\paragraph{Step 3: Projection and Visibility.} For each frame $f_t$, we project the 3D world point to 2D image coordinates using the projection operator $\Pi$:
\begin{equation}
    \vect{p}[f_t] = \Pi(\mat{K}, \mat{T}[f_t], \vect{X}[f_t])
\end{equation}
To determine visibility, we perform occlusion checking using depth comparison. We sample the observed depth map at the projected location using bilinear interpolation: $\hat{z}_t = \mathbf{D}[f_t](\vect{p}[f_t])$. The point is classified as visible if three conditions are satisfied: (1) the point projects within frame bounds; (2) the point is in front of the camera; and (3) the point is not occluded: $|z_t - \hat{z}_t| < \epsilon \cdot z_t$, where $z_t$ is the depth of $\vect{X}[f_t]$ in camera $f_t$, and $\epsilon = 0.1$ is the visibility threshold. Together, these steps yield a single 3D trajectory $\{\vect{X}[f_t]\}$ with per-frame visibility.

On top of this unified 3D trajectory, we define two complementary track representations. \emph{Visual point tracking} uses the time-varying camera poses to show how the point appears in each observed frame ($P_{2D}$), while \emph{true-motion point tracking} re-projects the trajectory into a fixed reference view to isolate object motion from camera motion ($M_{2D}$).

\paragraph{(12) Visual point tracking.}
Visual point tracking asks the model to predict $P_{2D} = \{\vect{p}[f_t]\}$. Given a query point $\vect{p}[f_q]$, we generate its full trajectory and create questions in two formats:
In the \emph{single-target} format, the question asks for the location of the point $\vect{p}[f_t]$ in one specific target frame $f_t$, and the answer is either its 2D coordinates or an ``occluded'' label. In the \emph{full-trajectory} format, the question instead asks for the entire sequence.

\begin{tcolorbox}[colback=green!5,colframe=green!40!black,title=\textbf{Example Question Template: Single-Target Visual Tracking}]
\small
\textbf{Question:} \textit{Given the red point at (0.451, 0.328) in frame 0, where is this point in frame 5?}

\vspace{2mm}
\textbf{Example Answers:}
\begin{itemize}[leftmargin=*, itemsep=1pt]
\item (0.804, 0.384)
\item The point is occluded in frame 5.
\end{itemize}
\end{tcolorbox}

\begin{tcolorbox}[colback=green!5,colframe=green!40!black,title=\textbf{Example Question Template: Full-Trajectory Visual Tracking}]
\small
\textbf{Question:} \textit{Given the red point at (0.451, 0.328) in frame 0, what is its trajectory across all frames?}

\vspace{2mm}
\textbf{Example Answer:}
\begin{itemize}[leftmargin=*, itemsep=1pt]
\item Frame 0: (0.451, 0.328), Frame 1: occluded, Frame 2: (0.486, 0.354), Frame 3: (0.522, 0.384)
\end{itemize}
\end{tcolorbox}

\paragraph{(13) True-motion point tracking.}
True-motion point tracking projects the 3D trajectory $\{\vect{X}[f_t]\}$ onto a fixed reference camera view at time $f_q$, isolating object motion from camera motion. This corresponds to computing $M_{2D} = \{\vect{m}_{f_q}[f_t]\}$:
\begin{equation}
    \vect{m}_{f_q}[f_t] = \Pi(\mat{K}, \mat{T}[f_q], \vect{X}[f_t])
\end{equation}
This produces a trajectory that shows how the point moves in a fixed camera reference frame. Questions follow the same format as visual point tracking but explicitly state the reference frame.

\begin{tcolorbox}[colback=green!5,colframe=green!40!black,title=\textbf{Example Question Template: True-Motion Tracking}]
\small
\textbf{Question:} \textit{What does the 3D trajectory of the blue point at (0.386, 0.414) (frame 2) look like when projected to frame 2?}

\vspace{2mm}
\textbf{Example Answer:} 
\begin{itemize}[leftmargin=*, itemsep=1pt]
\item Frame 0: (0.421, 0.454), Frame 1: (0.402, 0.436), Frame 2: (0.386, 0.414), Frame 3: (0.359, 0.386)
\end{itemize}
\end{tcolorbox}

\begin{table*}[t]
    \centering
    \vspace{-1em}
    \caption{
        \textbf{Additional benchmark results.} We evaluate the performance of the models trained on \datasetName on additional benchmarks.
    }
    \label{tab:additional_benchmarks}
    \resizebox{\linewidth}{!}{
    \begin{tabular}{l cc ccc cc}
        \toprule
        Methods
        & BLINK~\cite{fu2024blink}
        & MMSI~\cite{yang2025mmsi}
        & OmniSpatial~\cite{jia2025omnispatial}
        & SAT-Test~\cite{ray2024sat}
        & SAT-Val~\cite{ray2024sat}
        & VSTI~\cite{fan2025vlm}
        & SPAR~\cite{zhang2025flatland}
        \\
        \midrule
        \midrule
        Qwen2.5-VL-3B& 50.9 & 26.6 & 41.8 & 64.0 & 54.2 & 41.3 & 34.0
        \\
        \rowcolor{gray!20}& 49.3 & 26.9 & 40.4 & 68.7 & 68.0 & 39.1 & 37.7
        \\
        \rowcolor{gray!20} \multirow{-2}{*}{Qwen2.5-VL-3B + \datasetName} & 
        \textcolor{red}{-1.6} & \textcolor{green!60!black}{+0.3} & \textcolor{red}{-1.4} & \textcolor{green!60!black}{+4.7} & \textcolor{green!60!black}{+13.8} & \textcolor{red}{-2.2} & \textcolor{green!60!black}{+3.7}
        \\
        \midrule
        Qwen2.5-VL-7B& 52.5 & 24.2 & 37.7 & 46.7 & 57.0 & 25.9 & 41.7
        \\
        \rowcolor{gray!20}& 57.8 & 26.2 & 42.7 & 74.0 & 66.0 & 42.6 & 46.0
        \\
        \rowcolor{gray!20} \multirow{-2}{*}{Qwen2.5-VL-7B + \datasetName} & 
        \textcolor{green!60!black}{+5.3} & \textcolor{green!60!black}{+2.0} & \textcolor{green!60!black}{+5.0} & \textcolor{green!60!black}{+27.3} & \textcolor{green!60!black}{+9.0} & \textcolor{green!60!black}{+16.7} & \textcolor{green!60!black}{+4.3}
        \\
        \bottomrule
    \end{tabular}
    }
    \vspace{-1em}
\end{table*}

\subsection{Additional Answer Generation Details}
\label{supp:sec:answer_generation}

\subsubsection{Distractor Generation for Multiple-Choice Questions} For multiple-choice questions, we generate distractors (incorrect choices) tailored to each question type:

\vspace{-1em}\paragraph{(1) Motion questions:} We generate distractors by creating plausible but incorrect combinations of motion states. For example, if the correct answer is ``moving forward and up," distractors might be ``moving backward and up" or ``moving forward and down." We use a two-tier prioritization: \textit{mixed distractors} (containing both correct and incorrect motion components) are preferred over \textit{fully false distractors} (containing only incorrect components), as mixed distractors are more challenging.

\vspace{-1em}\paragraph{(2) Comparison questions:} For depth and distance comparisons, distractors are simply the alternative choices (e.g., if A is closer, then B is the distractor).

\subsubsection{Negative Question Augmentation} To improve robustness and test whether models can recognize non-existent objects, we augment a fraction of questions with \textit{negative questions}. With probability $p_{\text{neg}} = 0.1$, we replace one or more object references in the question with fabricated descriptions (e.g., ``the purple elephant" when no such object exists). For multiple-choice questions, the correct answer becomes a template like ``There is no such object in the video" or ``The object does not exist." For positive questions (where the object exists), we occasionally add ``There is no such object" as a distractor with the same probability $p_{\text{neg}}$, ensuring the model learns to correctly reject this choice when the object is present.

\section{Dataset Statistics}

\begin{figure}[t]
    \centering
    \includegraphics[width=\columnwidth]{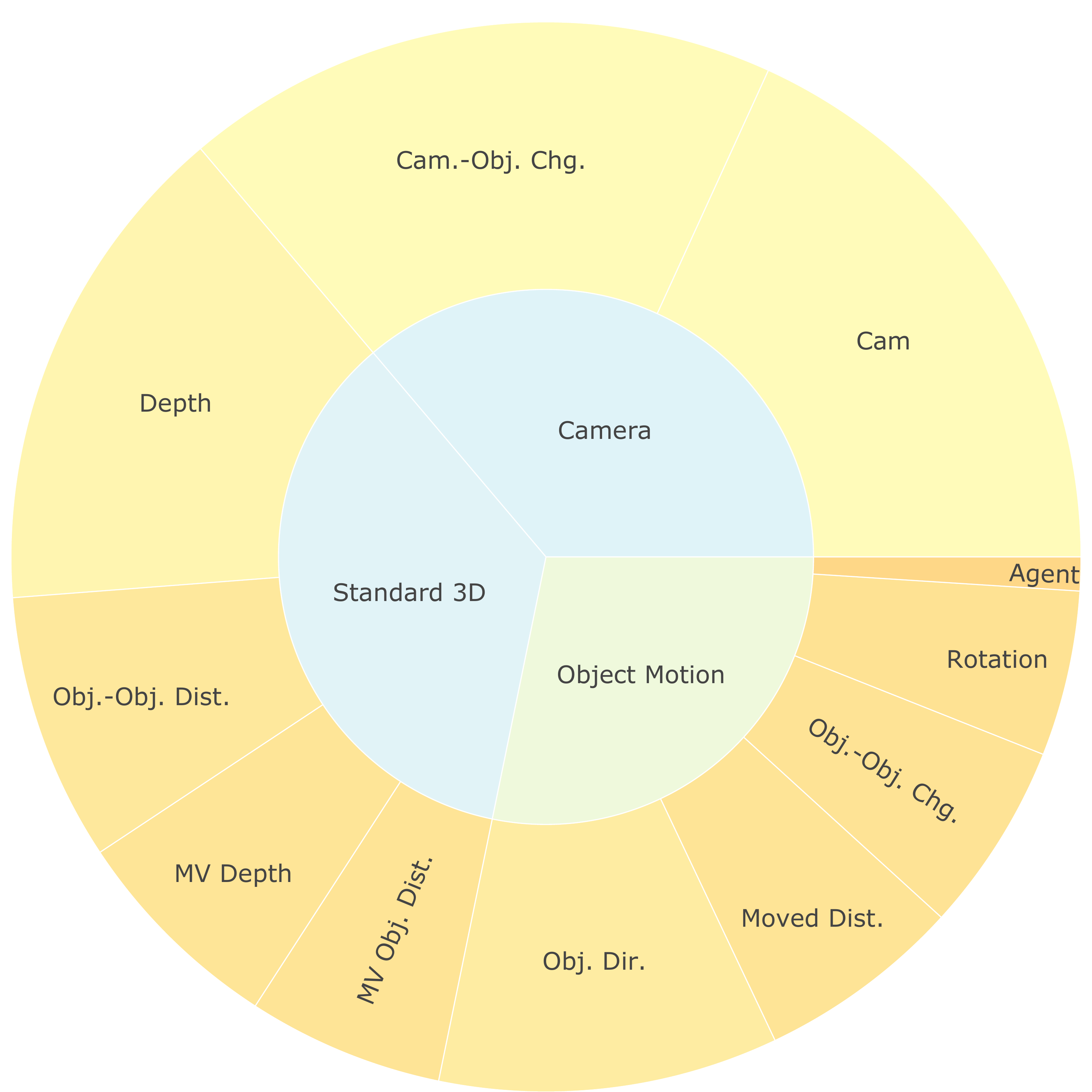}
    \caption{\textbf{Dataset statistics of \datasetName-Bench.}
    The dataset contains an even distribution of the three question categories: camera motion, object motion, and 3D spatial understanding.}
    \label{fig:statistic}
\end{figure}

\subsection{Question Type Distribution}
Figure~\ref{fig:statistic} illustrates the distribution of question types in \datasetName-Bench. The dataset is balanced across the three question types: camera motion, object motion, and 3D spatial understanding.

\subsection{Answer Format Distribution} Our training set consists of 400K QA pairs, approximately 20\% of which use multiple-choice format with 2-4 options, while the remaining 80\% are open-ended questions requiring coordinate outputs (point tracking) or per-frame classifications (detailed distance analysis). Negative questions comprise approximately 10\% of multiple-choice questions, ensuring models learn to recognize non-existent objects.

\section{Additional Results}

\subsection{Additional Benchmark Results}
\label{supp:sec:additional_benchmarks}

To further assess the generalization of \datasetName, we evaluate a baseline VLM fine-tuned solely on \datasetName across a diverse suite of established benchmarks (Table~\ref{tab:additional_benchmarks}).
Specifically, BLINK~\cite{fu2024blink} evaluates a broad range of general visual perceptual abilities.
OmniSpatial~\cite{jia2025omnispatial} and MMSI~\cite{yang2025mmsi} assess comprehensive spatial reasoning across diverse question types.
SAT~\cite{ray2024sat} and VSTI~\cite{fan2025vlm} cover both static and dynamic spatial scene understanding in video.
SPAR~\cite{zhang2025flatland} evaluates 3D spatial perception and reasoning from 2D inputs.
Achieving strong performance across this broad set of benchmarks generally benefits from training on dataset mixtures combining task-specific data sources~\cite{liu2024nvila,bai2025qwen2,yang2025qwen3}, which we leave as future work.
Nonetheless, we report results of fine-tuning exclusively on \datasetName to demonstrate its generalization without any task-specific mixture training.
We observe consistent performance improvements for Qwen2.5-VL-7B across all benchmarks, demonstrating that the low-level 4D understanding signals learned from \datasetName transfer to a wide range of spatial and perceptual tasks.
Importantly, this generalizability is achieved without exposure to any of these target benchmarks during fine-tuning.

\subsection{QA-Pair Bias Analysis}
\label{supp:sec:bias_analysis}

To assess the quality of our QA pairs and investigate potential template biases inherent to synthetically constructed datasets, we evaluate a set of carefully designed baselines on \datasetName-Bench.
Results are summarized in Table~\ref{tab:random_blind_ablation}.

Since our benchmark covers both multiple-choice and yes/no question formats, the expected random-choice accuracy is $40.8\%$.
To probe whether language priors or template-level cues alone can be exploited without visual information, we evaluate two blind baselines by replacing all video frames with fully blacked-out images.
First, the proprietary Gemini-2.5-Pro achieves $40.0\%$ accuracy under blind inputs, closely matching the random baseline.
This indicates that language priors or template-level cues alone are insufficient to solve our benchmark.
Second, we train Qwen2.5-VL-7B on blind input videos, allowing the model to potentially exploit any dataset-specific distributional biases or shortcuts.
This variant achieves $52.5\%$ accuracy, revealing a limited but measurable distributional bias.
Such bias likely arises from inherent tendencies present in the underlying data sources used to construct our dataset.

In contrast, Qwen2.5-VL-7B with full visual input reaches $85.1\%$ accuracy, confirming that visual reasoning is the primary driver of performance.
While the model can exploit minor linguistic biases and correlations, as commonly observed in VQA datasets~\cite{brown2025benchmark}, the substantial gain with visual input ($+32.6\%$) underscores that language priors alone cannot solve the task.

\begin{table}[t]
    \centering
    \caption{\textbf{QA-pair bias analysis on \datasetName-Bench.}
    Blind baselines replace all video frames with fully blacked-out images.
    Performance close to the random baseline ($40.8\%$) confirms that language priors alone cannot solve the task, while the large gain from full visual input demonstrates that visual reasoning drives performance.}
    \label{tab:random_blind_ablation}
    \resizebox{\linewidth}{!}{%
    \begin{tabular}{@{}l c c c c@{}}
    \toprule
    \textbf{Model} & \textbf{Camera Motion} & \textbf{Object Motion} & \textbf{3D Spatial} & \textbf{Overall} \\
    \midrule
    Random & 41.5 & 28.5 & 50.0 & 40.8 \\
    Gemini-2.5-Pro (blind) & 37.7 & 44.7 &  40.1 & 40.0 \\
    \midrule
    NVILA-Lite-8B & 42.4 & 26.0 & 55.4 & 42.3 \\
    NVILA-Lite-8B + \datasetName~(blind) & 55.5 & 45.1 & 63.6 & 55.1 \\
    NVILA-Lite-8B + \datasetName & 83.5 & 81.6 & 88.6 & 84.4 \\
    \arrayrulecolor{black}
    \bottomrule
    \end{tabular}%
    }
\end{table}

\subsection{Quantitative Results of Tracking Tasks}

\begin{table}[t]
    \centering
    \caption{\textbf{Quantitative results of tracking tasks.} 
    We evaluate the performance of the model on the point tracking tasks, namely, visual point tracking (VPT) and true-motion point tracking (TMPT). We use average Jaccard (AJ) score~\cite{doersch2022tap} to evaluate the performance of the model on the tracking tasks.
    }
    \resizebox{\linewidth}{!}{
    \begin{tabular}{lcc}
    \toprule
    \multirow{2}{*}{Method} & \multicolumn{2}{c}{Tracking Task} \\
    \cmidrule(lr){2-3}
    & VPT & TMPT  \\
    \midrule
    NVILA-Lite-8B + Std-\datasetName + VPT & 43.2 & --  \\
    NVILA-Lite-8B + Std-\datasetName + TMPT & -- & 51.0 \\
    NVILA-Lite-8B + Std-\datasetName + VPT + TMPT & 44.1 & 48.8 \\
    \bottomrule
    \end{tabular}
    }
    \label{tab:tracking_results}
\end{table}

In this experiment, we extract 200 tracking instances, with 100 cases of visual point tracking (VPT) and 100 cases of true-motion point tracking (TMPT). We then evaluate model performance on these tracking tasks using the average Jaccard (AJ) score~\cite{doersch2022tap}, which measures both the accuracy of the predicted point locations and the occlusion predictions.
The results are shown in Table~\ref{tab:tracking_results}.

\subsection{Visualization of the \datasetName-Bench}
\label{supp:sec:visualization}
We include an interactive \datasetName-Bench visualizer in the supplementary material to facilitate a better understanding of the question and answer distribution. This tool displays QA pairs from each dataset: ADT~\cite{pan2023aria}, HOT3D~\cite{banerjee2024introducing}, Virtual KITTI2~\cite{cabon2020virtual}, Kubric~\cite{greff2022kubric}, and SHIFT~\cite{sun2022shift}. Due to space constraints, we present a subset of 212 QA pairs. To access the visualization, please open the \texttt{index.html} file in a web browser.

\section{Limitations}
We highlight some of the limitations of the \datasetName-Bench and the \datasetName dataset. Our current framework focuses primarily on rigid object motion, potentially limiting its applicability to complex non-rigid dynamics or highly deformable objects. Moreover, the use of hysteresis thresholding to filter out ambiguous motion helps ensure annotation reliability but also means that the dataset does not cover subtle motion cases that fall between the defined thresholds. We believe that these limitations can be addressed by extending the framework to cover more complex and subtle motion cases, and we leave this as future work.

\clearpage
{
    \small
    \bibliographystyle{ieeenat_fullname}
    \bibliography{main}
}

\end{document}